\documentclass[10pt,twocolumn,letterpaper]{article}
\usepackage[pagenumbers]{cvpr}     

\definecolor{cvprblue}{rgb}{0.21,0.49,0.74}
\usepackage[pagebackref,breaklinks,colorlinks,allcolors=cvprblue,urlcolor=magenta]{hyperref}
\usepackage{multirow}
\usepackage[table]{xcolor}
\usepackage{makecell}
\usepackage{verbatim}
\usepackage[accsupp]{axessibility}

\def\confName{CVPR}
\def\confYear{2026}

\title{A Semantically Disentangled Unified Model \\for Multi-category 3D Anomaly Detection}

\author{
SuYeon Kim \quad
Wongyu Lee \quad
MyeongAh Cho$^{\dagger}$ \\
Kyung Hee University, Republic of Korea\\
\small\texttt{\{spoiuy3, fovert, maycho\}@khu.ac.kr}
}

\begin{document}
\twocolumn[{
\renewcommand\twocolumn[1][]{#1}
\maketitle
\begin{center}
    \centering
    \captionsetup{type=figure}\vspace{-5mm}
    \includegraphics[width=\textwidth]
    {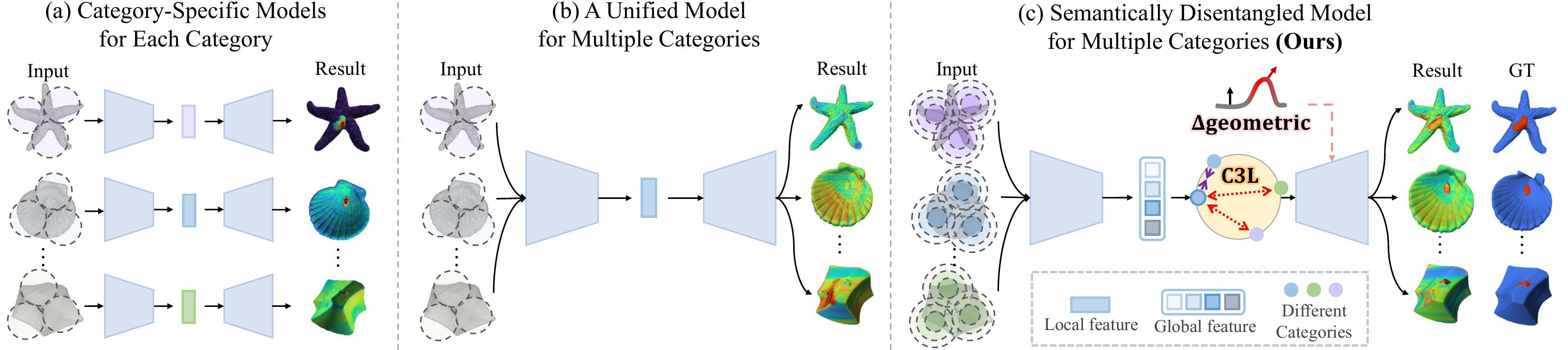}
    \captionof{figure}{Overview of different paradigms for 3D anomaly detection.
(a) Category-specific models trained on each object class separately.
(b) A unified model handles multiple categories with local features but often suffers from inter-category feature entanglement.
(c) Our method achieves semantically consistent results by aggregating coarse-to-fine geometric cues into category-aware global features, disentangling them via C3L, and guiding reconstruction with input geometry.}
    \label{fig:figure1}
\end{center}
}]

\footnotetext{\dag\ Corresponding author}
\newcommand*{\rom}[1]
{\expandafter\@slowromancap\romannumeral #1@}
\begin{abstract}
3D anomaly detection targets the detection and localization of defects in 3D point clouds trained solely on normal data.
While a unified model improves scalability by learning across multiple categories, it often suffers from \textbf{Inter-Category Entanglement (ICE)}—where latent features from different categories overlap, causing the model to adopt incorrect semantic priors during reconstruction and ultimately yielding unreliable anomaly scores.
To address this issue, we propose the \textbf{Semantically Disentangled Unified Model} for 3D Anomaly Detection, which reconstructs features conditioned on disentangled semantic representations.
Our framework consists of three key components: (i) \textbf{Coarse-to-Fine Global Tokenization} for forming instance-level semantic identity, (ii) \textbf{Category-Conditioned Contrastive Learning} for disentangling category semantics, and (iii) a \textbf{Geometry-Guided Decoder} for semantically consistent reconstruction.
Extensive experiments on Real3D-AD and Anomaly-ShapeNet demonstrate that our method achieves state-of-the-art for both unified and category-specific models, improving object-level AUROC by 2.8\% and 9.1\%, respectively, while enhancing the reliability of unified 3D anomaly detection. Project page: \small\textcolor{magenta}{\href{https://visualsciencelab-khu.github.io/SeDiR_project/}{\texttt{https://spoiuy3.github.io/SeDiR/}}}
\end{abstract}    
\vspace{-3mm}
\section{Introduction}
\label{sec:intro}
\setlength{\abovecaptionskip}{5pt}
\setlength{\belowcaptionskip}{5pt}

3D anomaly detection (3D-AD) aims to identify defective objects and localize abnormal regions within 3D point clouds, typically trained using only normal data. Conventional methods~\cite{liu2023real3d, zhu2024towards, li2024towards, zhou2024r3d} are category-specific, where a separate model is trained for each object type (\cref{fig:figure1} (a)). These models—most commonly based on encoder–decoder reconstruction or memory matching—learn the normal pattern distribution of a single category. During inference, reconstruction-based methods detect anomalies when the model fails to reconstruct an abnormal sample, whereas memory-based approaches identify anomalies when the feature representation of an abnormal sample deviates significantly from the stored normal patterns. Although effective under controlled conditions, this paradigm has two major limitations. First, it suffers from \textbf{transfer collapse}—the model becomes overly specialized to a specific geometric pattern of normal training samples and fails to generalize over shape or texture variations. Second, it is \textbf{impractical in real industrial scenarios} involving multiple object categories, as it requires duplicating training, storage, and maintenance pipelines for each class. To address these constraints, a recent study~\cite{ijcai2025p94} has explored unified 3D-AD framework, where a single model is jointly trained across multiple categories (\cref{fig:figure1} (b)). Such unified design reduces system redundancy and improves deployment efficiency, emerging as a promising direction for practical 3D anomaly detection.

A unified model offers improved practicality and even higher accuracy, yet a critical question remains—\textbf{can a model truly detect anomalies without first establishing an object’s identity?} 
In unified 3D-AD, the model is expected to learn normal shape priors across multiple categories and detect anomalies through reconstruction errors. 
However, our observations reveal a deeper issue: even when trained only on normal data, the model sometimes reconstructs objects using the wrong category prior (\eg, a chair partially reconstructed with table-like geometry). This does not indicate a failure to detect anomalies, but rather a failure to establish the object’s identity before reconstruction. We define this phenomenon as \textbf{Inter-Category Entanglement (\textit{ICE})}—a state in which features from different categories become entangled in the latent space, preventing the formation of distinct, category-specific representations. When the \textit{ICE} problem occurs, the model loses semantic identity, and reconstruction is guided by an incorrect category prior. This semantic misalignment propagates through the entire pipeline, leading to geometrically inconsistent outputs and degraded performance in both anomaly detection and localization.

\begin{figure}[!t] 
    \centerline{\includegraphics[width=\linewidth]{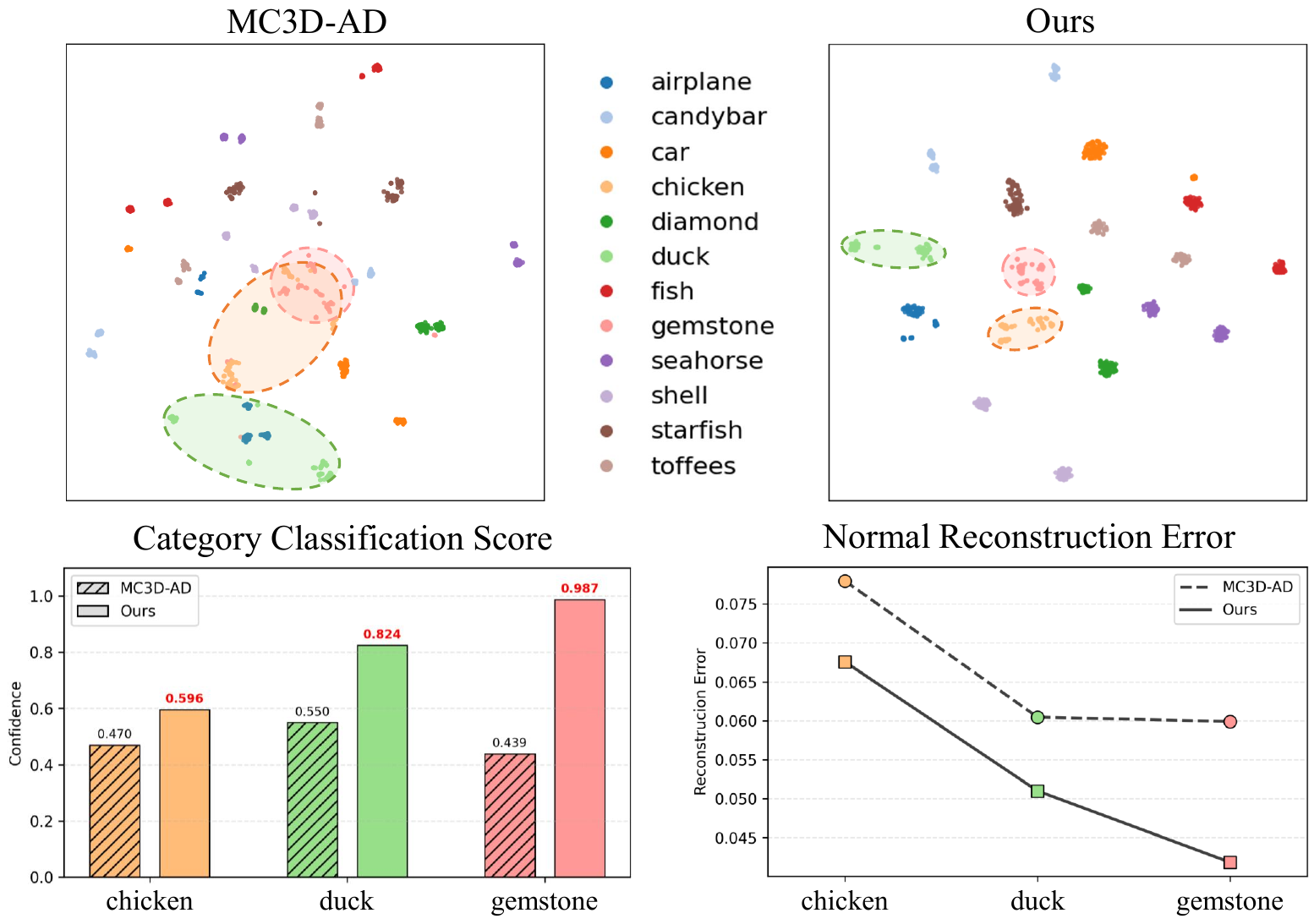}}
    \caption{T-SNE visualization and quantitative analysis of semantic disentanglement. MC3D-AD exhibits entangled feature clusters across similar categories (\eg, \textit{chicken, duck, gemstone}), where samples with low category classification scores are reconstructed under uncertain semantic priors, yielding high reconstruction errors on normal data and increased false positives. Our model instead forms well-separated, semantically aligned manifolds with higher classification scores and lower reconstruction errors, enabling reliable category-conditioned reconstruction.}
    \label{fig:figure2} 
    \vspace{-5mm}
\end{figure}

\begin{figure*}[!t]
    \centerline{\includegraphics[width=\linewidth]{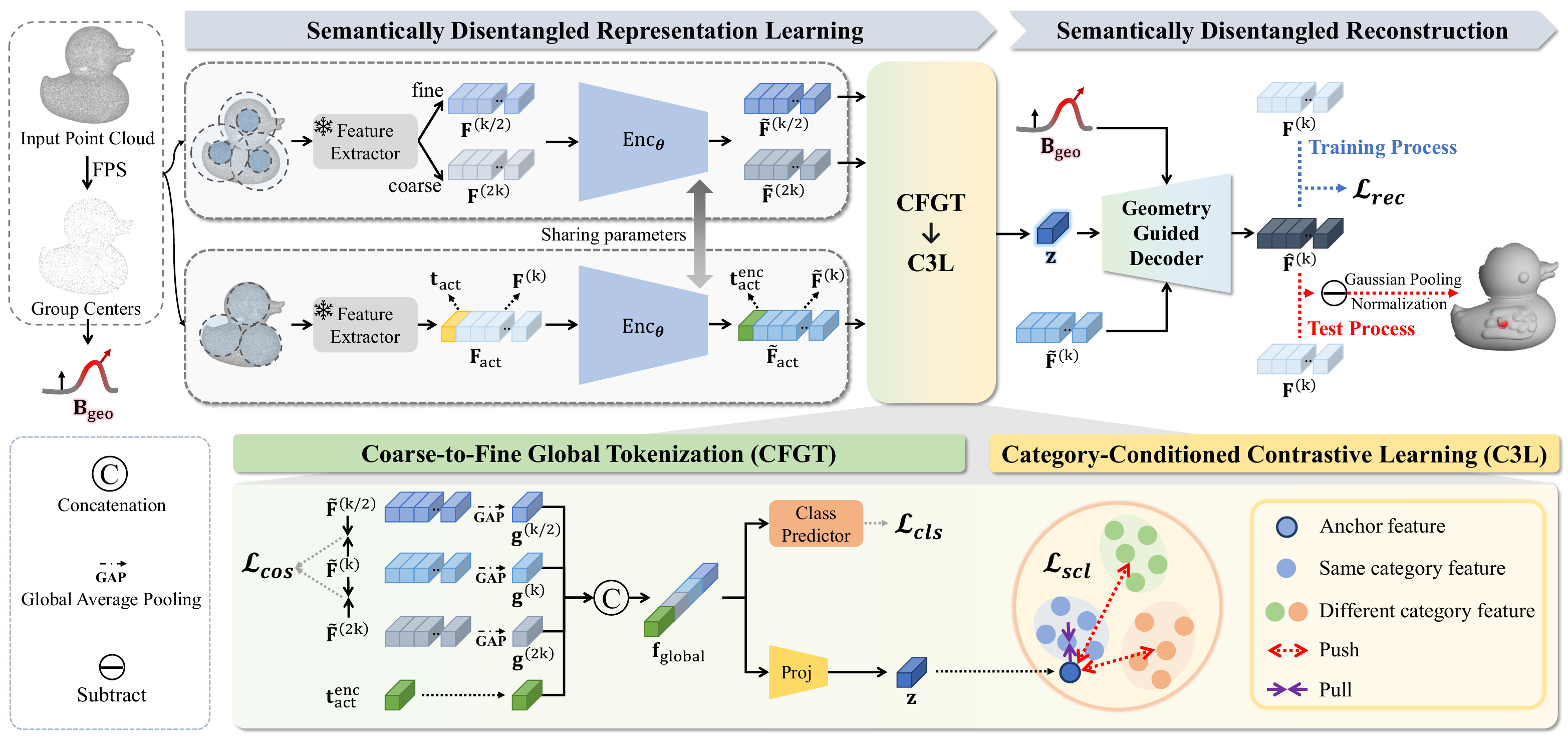}}
    \caption{\textbf{The overview of the proposed method.} Our method consists of two main stages: Semantically Disentangled Representation Learning and Semantically Disentangled Reconstruction. 
    Given an input point cloud, CFGT encodes multi-resolution geometric features into a category-aware global token, C3L disentangles the latent semantics, and GGD reconstructs the object conditioned on these disentangled semantics and geometric priors.}
    \label{fig:figure_3} 
    \vspace{-5mm}
\end{figure*}

To validate the presence of the \textit{ICE} problem, in \cref{fig:figure2}, we perform both qualitative and quantitative analyses using MC3D-AD~\cite{ijcai2025p94}. We first project encoder features into t-SNE space~\cite{maaten2008visualizing} and observe that categories such as \textit{chicken, duck,} and \textit{gemstone} form heavily overlapping clusters rather than distinct manifolds. 
Furthermore, to quantify the category classification score, we freeze the trained model and train a lightweight MLP on the reconstructed features.
By analyzing the relationship between the classification score and the reconstruction error on normal test samples, we observe that samples with low semantic score tend to exhibit significantly higher reconstruction errors. These observations confirm that incorrect semantic understanding leads to incorrect reconstruction, demonstrating that the \textit{ICE} problem is not a random side effect but a systematic bottleneck in unified 3D-AD.

The above observations indicate that the fundamental limitation of unified 3D-AD is not the difficulty of reconstructing anomalies, but the attempt to reconstruct before understanding what is being reconstructed. In other words, reconstruction fails not because the object is anomalous, but because the model lacks a clear semantic identity of the object in the first place. This misalignment—where geometry is reconstructed under an incorrect category prior—leads to unreliable anomaly scores and mislocalized defects. Therefore, we reformulate unified 3D-AD as a problem of \textbf{semantically conditioned reconstruction} and propose the paradigm of \textbf{Semantically Disentangled Reconstruction} (\cref{fig:figure1} (c)), where reconstruction is conditioned on correctly disentangled, category-level semantics rather than indiscriminately applied to all latent representations.

To operationalize Semantically Disentangled Reconstruction, we design a semantically disentangled unified framework, named \textbf{SeDiR}, that explicitly separates semantic identity from geometry before attempting reconstruction. We propose the \textbf{Coarse-to-Fine Global Tokenization (CFGT)} module, which aggregates multi-resolution geometric cues into a global token that captures category-aware semantics rather than relying solely on local point features. This global representation enables the model to form a reliable semantic identity of the object before any decoding takes place. To further prevent different categories from collapsing into a shared latent space, we introduce \textbf{Category-Conditioned Contrastive Learning (C3L)}, which explicitly enforces intra-category compactness and inter-category separation to form distinct semantic manifolds, directly mitigating Inter-Category Entanglement. Finally, we propose a \textbf{Geometry-Guided Decoder (GGD)} that performs reconstruction guided by both the disentangled semantic prior and local geometric evidence. Instead of decoding blindly from latent features, the decoder is softly biased towards category-consistent reconstruction pathways, ensuring that geometry is reconstructed according to the correct semantic identity. Together, these components enable the model to understand \textit{what} it is reconstructing before deciding \textit{how} to reconstruct, establishing a new perspective and solution space for unified 3D anomaly detection.
Extensive experiments validate the effectiveness of our framework, showing consistent improvements and outperforming both category-specific and unified state-of-the-art baselines by 2.8\% and 9.1\% in object-level AUROC on Real3D-AD and Anomaly-ShapeNet, respectively.
\section{Related Works}
\label{sec:related_works}

\subsection{2D Anomaly Detection}
Unsupervised 2D image-based anomaly detection (2D-AD) has advanced rapidly in recent years, mainly through two paradigms: feature-embedding and reconstruction-based methods.
Feature-embedding methods model the latent distribution of normal data using pretrained representations and detect anomalies by measuring the distance from the normal feature manifold~\cite{yi2020patch, defard2021padim, roth2022towards, strater2024generalad, wang2025distribution}.
Reconstruction-based approaches train generative models such as autoencoders~\cite{zavrtanik2021reconstruction, ristea2022self, huang2022self, tao2022deep, liu2025simple}, GANs~\cite{yan2021learning, liang2023omni, fang2025boosting}, transformers~\cite{pirnay2022inpainting, yao2024prior, iqbal2024multi}, or diffusion models~\cite{wyatt2022anoddpm, yao2024glad, fuvcka2024transfusion, he2024diffusion} to reconstruct normal images and use the reconstruction error as the anomaly score.
Recently, unified frameworks have been proposed to replace class-specific models, enabling generalization across categories~\cite{you2022unified, zhao2023omnial, lu2023hierarchical, he2024mambaad, yao2024resad, guo2025dinomaly}.
Some of these methods~\cite{wei2025uninet, fan2025salvaging} employ contrastive learning to structure feature spaces with clearer inter-class boundaries.
However, these approaches focus solely on semantic separation in 2D feature space and lack explicit mechanisms for geometric consistency, making direct extension to 3D data difficult.

\subsection{3D Anomaly Detection}
Research on 3D-AD with point clouds has accelerated in recent years. Existing approaches are broadly grouped into memory bank–based and reconstruction-based methods. 
Memory bank–based methods store features of normal point clouds and compare test samples using similarity metrics~\cite{liu2023real3d, zhu2024towards, liang2025look}.
Reconstruction-based methods instead learn to reproduce normal shapes and identify anomalies through reconstruction errors~\cite{li2024towards, zhou2024r3d}.
Similar to 2D-AD, unified 3D-AD framework has recently emerged to enable anomaly detection across multiple categories within a single model.
MC3D-AD~\cite{ijcai2025p94} aims to generalize across categories but often suffers from \textit{inter-category entanglement (ICE)}, where features from different classes interfere with one another.
This highlights the need for disentangling category-specific semantics while preserving geometric consistency—a challenge our method directly addresses.

\section{Method}
\label{sec:method}

\noindent\textbf{Problem statement.}
Let $\mathcal{C}=\{1,\ldots,C\}$ be the set of object categories.
For each category $c \in \mathcal{C}$, we denote the training set by $\mathcal{D}^{\,c}_{\text{train}}=\{\,P^{c}_{q}\,\}_{q=1}^{M_c}$,
where each point cloud $P^{c}_{q}\in\mathbb{R}^{N\times 3}$ contains $N$ points, and $M_c$ is the number of training samples.
The corresponding test set is given by $\mathcal{D}^{\,c}_{\text{test}}=\{\,\big(P^{c}_{q},\, t^{c}_{q}\big)\,\}_{q=1}^{T_c}$,
where $t^{c}_{q}\in\mathcal{T}:=\{0,1\}$ indicates whether the sample is normal ($0$) or anomalous ($1$).
For the multi-category 3D anomaly detection setting,
we construct unified datasets across all categories as
$\mathcal{D}_{\text{train}}=\bigcup_{c=1}^{C}\mathcal{D}^{\,c}_{\text{train}}$ and $\mathcal{D}_{\text{test}}=\bigcup_{c=1}^{C}\mathcal{D}^{\,c}_{\text{test}}$.
The goal of this task is to train a unified model using only normal samples from $\mathcal{D}_{\text{train}}$, and evaluate it on  object-level anomaly detection and point-level anomaly localization.

\noindent\textbf{Overview.}
The overview of our framework \textbf{SeDiR} is illustrated in \cref{fig:figure_3}.
Given an input point cloud, multi-resolution geometric features are extracted via a shared encoder, where a learnable token $\mathbf{t}_{\mathrm{act}}$ is appended to capture instance-level contextual information.
These feature sequences are processed by the CFGT to generate a category-aware global token and are further optimized with C3L to achieve semantically disentangled feature separation.
The GGD then reconstructs feature tokens in a semantically consistent manner, conditioned on both the global token and geometric bias guided by the input geometry.
During inference, reconstruction errors are aggregated using Gaussian pooling and normalization to compute the final anomaly score.
The details of each component are described in the following sections.
\subsection{Coarse-to-Fine Global Tokenization}
\label{sec:cfgf}

\noindent\textbf{Multi-resolution neighborhood encoding.}
Given a point cloud \(\mathbf{P}=\{\mathbf{p}_n\}_{n=1}^{N}\), \(\mathbf{p}_n\in\mathbb{R}^{3}\), 
we first select \(g\) centers \(\mathcal{S}=\{\mathbf{s}_{m}\}_{m=1}^{g}\) via Farthest Point Sampling (FPS)~\cite{qi2017pointnet++}.
Based on a base neighborhood size $k$, we define a set of symmetric resolutions 
$\mathcal{R}=\{k/2,\,k,\,2k\}$ to capture both fine- and coarse-scale geometric context.
For each \(r\in\mathcal{R}\), Euclidean neighborhoods are constructed around the shared centers \(\mathcal{S}\) using the \(k\)-nearest neighbors (kNN) algorithm, 
and each neighborhood is encoded using a pretrained local encoder \(E(\cdot)\) (\eg, PointMAE~\cite{pang2023masked}):
\begin{equation}
\mathcal{N}_{r}(\mathbf{s}_m)=\mathrm{kNN}(\mathbf{s}_m;\,r),\quad
\mathbf{f}^{(r)}_{m}=E\!\big(\mathcal{N}_{r}(\mathbf{s}_m)\big)\in\mathbb{R}^{d}.
\label{eq:knn_mra}
\end{equation}
Smaller neighborhoods (\(k/2\)) focus on fine details, while larger ones (\(2k\)) encode broader structural context.
Collectively, these multi-resolution neighborhoods capture complementary geometric cues across scales, forming a coarse-to-fine representation.
Finally, the resulting feature tokens across all centers are aggregated into a resolution-specific feature sequence:
\begin{equation}
\mathbf{F}^{(r)}=\big[\mathbf{f}^{(r)}_{1},\ldots,\mathbf{f}^{(r)}_{g}\big]\in\mathbb{R}^{g\times d}.
\label{eq:mra_feats}
\end{equation}
Additional experiments exploring various symmetric resolution configurations are included in the supplementary.

\noindent\textbf{Instance-level global representation via adaptive context token.} 
Although multi-resolution encoding captures geometry, it lacks an instance-level semantic representation.
To construct such a global representation, we introduce a learnable \emph{Adaptive Context Token} (ACT), denoted as \(\mathbf{t}_{\mathrm{act}}\in\mathbb{R}^{1\times d}\), and prepend it to the base-resolution feature sequence, forming
\(\mathbf{F}_{\mathrm{act}} = \operatorname{concat}([\mathbf{t}_{\mathrm{act}}, \mathbf{F}^{(k)}]) \in \mathbb{R}^{(g+1)\times d}\).
The ACT dynamically aggregates global context.
A transformer encoder $\texttt{Enc}_{\boldsymbol{\theta}}$ with shared weights processes both the coarse-to-fine sequences and the ACT-augmented base sequence:
\begin{equation}
\tilde{\mathbf{F}}^{(r)} = \texttt{Enc}_{\boldsymbol{\theta}}(\mathbf{F}^{(r)}), \quad
\tilde{\mathbf{F}}_{\mathrm{act}} = \texttt{Enc}_{\boldsymbol{\theta}}(\mathbf{F}_{\mathrm{act}}),
\end{equation}
where \(r\in\{k/2,\,2k\}\).
The output of \(\tilde{\mathbf{F}}_{\mathrm{act}}\) excluding the ACT corresponds to the base-resolution feature sequence \(\tilde{\mathbf{F}}^{(k)}\).

To encourage cross-scale consistency, we impose a cosine alignment loss between the base tokens and their fine- and coarse-scale counterparts:
\begin{equation}
\mathcal{L}_{\mathrm{cos}}
=\frac{1}{g}\sum_{m=1}^{g}\sum_{r\in\{k/2, 2k\}}
\Big[1-\cos\!\big(\tilde{\mathbf{f}}^{(k)}_{m},\,\tilde{\mathbf{f}}^{(r)}_{m}\big)\Big],
\label{eq:cos_align}
\end{equation}
where \(\cos(\cdot,\cdot)\) denotes the normalized dot product.

To facilitate cross-scale fusion, we summarize the contextual information of each feature sequence via global average pooling,
\(\mathbf{g}^{(r)} = \operatorname{GlobalAveragePool}(\tilde{\mathbf{F}}^{(r)})\),
where \(r\in\{k/2,\,k,\,2k\}\),
and directly extract the encoded ACT token as
\(\mathbf{t}^{\mathrm{enc}}_{\mathrm{act}} = \tilde{\mathbf{F}}_{\mathrm{act}}[1,:]\).
Finally, we obtain the holistic global representation by concatenating the pooled fine-, base-, and coarse-level features together with the ACT token:
\begin{equation}
\mathbf{f}_{\mathrm{global}}
= \operatorname{concat}([\mathbf{g}^{(k)},\,\mathbf{g}^{(2k)},\,\mathbf{g}^{(k/2)},\,\mathbf{t}^{\mathrm{enc}}_{\mathrm{act}}]),
\label{eq:global_concat}
\end{equation}
which serves as a unified global representation encapsulating both geometric structure and semantic identity.
Further analysis on ACT behavior is in the supplementary.

\noindent\textbf{Auxiliary category classification loss.}
To imbue the global embedding with category awareness, 
we introduce an auxiliary classification head that predicts the object category as 
\(\hat{\mathbf{y}}=\texttt{Cls}(\mathbf{f}_{\mathrm{global}})\).
Let \(\mathbf{y}\in\{0,1\}^{C}\) denote the one-hot label for category \(c\in\{1,\dots,C\}\).
The auxiliary objective is formulated as a standard cross-entropy loss:
\begin{equation}
\mathcal{L}_{\mathrm{cls}}=\operatorname{CrossEntropy}\!\big(\hat{\mathbf{y}},\,\mathbf{y}\big).
\label{eq:cls_loss}
\end{equation}
This auxiliary supervision promotes category-aware representations, facilitating semantic disentanglement.

\noindent\textbf{Global token embedding.}
For each instance, the aggregated global representation \(\mathbf{f}_{\text{global}}\)—which encodes both geometric structure and category-aware semantics—is projected through a lightweight head \(\texttt{Proj}(\cdot)\) and subsequently \(\ell_2\)-normalized to form a compact global token:
\begin{equation}
\mathbf{z} = \texttt{Proj}(\mathbf{f}_{\text{global}}), \quad
\mathbf{z} \in \mathbb{R}^{d_z}, \quad
\mathbf{z} = \frac{\mathbf{z}}{\lVert \mathbf{z} \rVert_2}.
\label{eq:projection}
\end{equation}
The projector is a two-layer MLP with LayerNorm~\cite{ba2016layer} and GeLU~\cite{hendrycks2016gaussian}, outputting an embedding of dimension \(d_z\).
The resulting normalized global token serves as a compact instance descriptor and is further utilized in \cref{subsec:C3L} and \cref{subsec:geom_decoder}.

\begin{figure}[!t]
    \centering
    \includegraphics[width=0.9\linewidth, height=5.0cm, keepaspectratio]{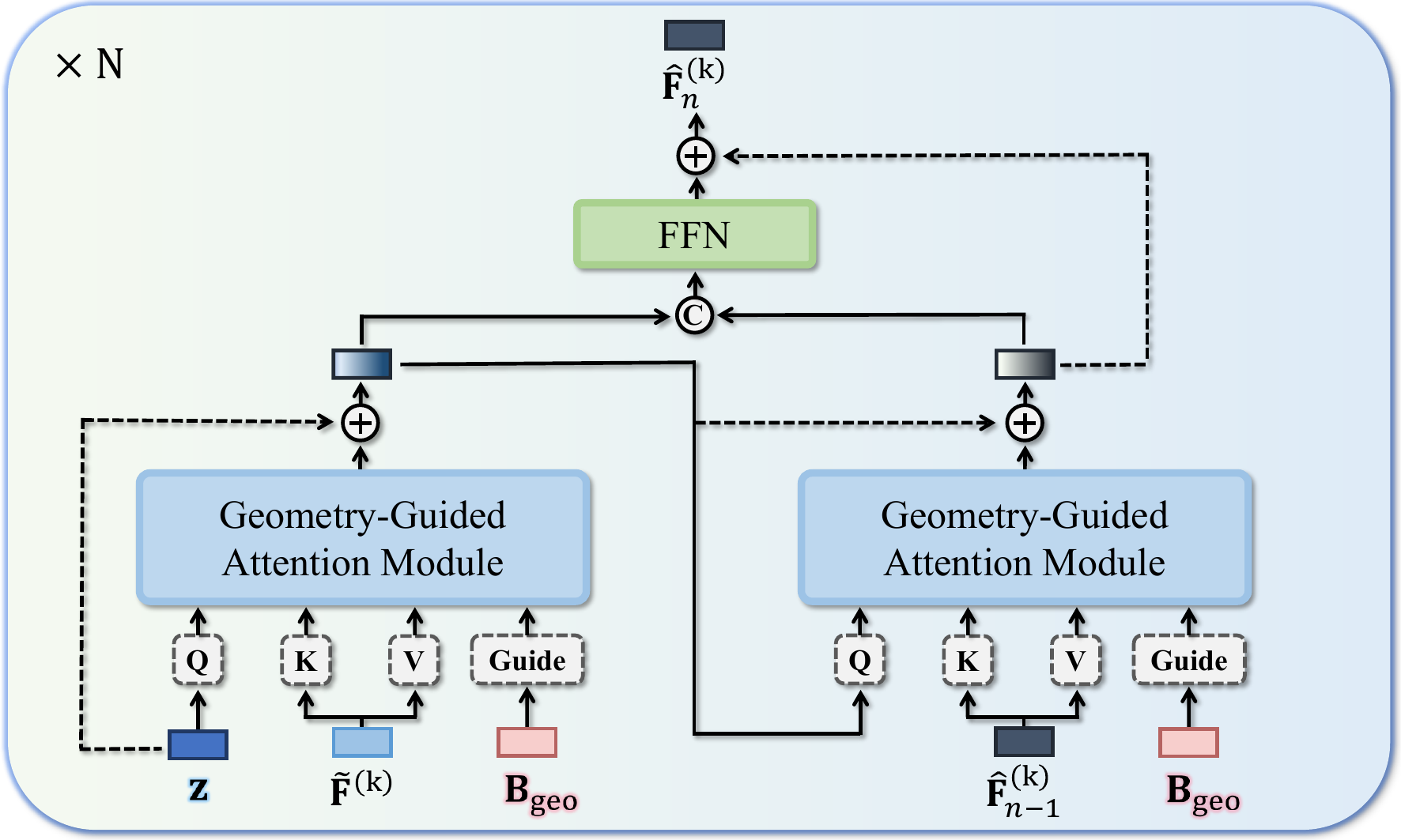} 
    \caption{Overview of the Geometry-Guided Decoder (GGD). Geometric priors $\mathbf{B}_{\mathrm{geo}}$ guide the attention mechanism to refine semantic features, and the results are aggregated through a feed-forward network.}
    \label{fig:figure_4} 
    \vspace{-5mm}
\end{figure}

\subsection{Category-Conditioned Contrastive Learning}
\label{subsec:C3L}

To mitigate \textit{inter-category entanglement (ICE)}, we propose Category-Conditioned Contrastive Learning (C3L), which further refines the latent space built upon the category-aware global token from CFGT.
We construct a dynamic buffer 
\(\mathcal{B}=\{(\mathbf{z}_j,\,c_j)\mid j=1,\dots,L\}\)
to ensure stable and diverse contrastive samples,
where \(c_j\in\{1,\dots,C\}\) denotes the category label.
We set L to 64, and additional analysis on buffer size sensitivity is provided in the supplementary.
For a given embedding \(\mathbf{z}_i\) with category \(c_i\),
we define its positive and negative sets as
\(\mathcal{P}(i)=\{\mathbf{z}_{pos}\in\mathcal{B}\mid c_p=c_i\}\),
\(\mathcal{N}(i)=\{\mathbf{z}_{neg}\in\mathcal{B}\mid c_n\neq c_i\}\),
and \(\mathcal{A}(i)=\mathcal{P}(i)\cup\mathcal{N}(i)\).
Following the supervised contrastive learning formulation~\cite{khosla2020supervised}, the per-sample supervised contrastive loss is formulated as:
{\small\begin{equation}
\mathcal{L}_{\mathrm{scl}}(i)=
\frac{1}{|\mathcal{P}(i)|}
\sum_{\mathbf{z}_{pos}\in\mathcal{P}(i)}
-\log
\frac{\exp(\mathbf{z}_i^\top\mathbf{z}_{pos}/\tau)}
{\sum\limits_{\mathbf{z}_a\in\mathcal{A}(i)}\exp(\mathbf{z}_i^\top\mathbf{z}_a/\tau)},
\label{eq:scl_single}
\end{equation}}
where \(\tau\) denotes the temperature parameter.

\noindent\textbf{Training objective.}
The final C3L objective combines the supervised contrastive loss, 
the auxiliary category classification term, and the cross-scale cosine alignment term:
\begin{equation}
\mathcal{L}_{\mathrm{C3L}}
=\lambda_\mathrm{scl}\mathcal{L}_{\mathrm{scl}}
+\lambda_\mathrm{cls}\mathcal{L}_{\mathrm{cls}}
+\lambda_\mathrm{cos}\mathcal{L}_{\mathrm{cos}},
\label{eq:C3L_loss}
\end{equation}
where \(\lambda_\mathrm{scl}\), \(\lambda_\mathrm{cls}\), and \(\lambda_\mathrm{cos}\) 
balance the respective objectives.
This joint optimization disentangles inter-category embeddings 
while preserving compact intra-category geometry, 
ultimately yielding a semantically disentangled and category-coherent latent space.

\begin{table*}[!ht]
\caption{Quantitative comparison of AUROC (\%) at the object levels on Real3D-AD across 12 categories. The best and second-best results are highlighted in \textbf{bold} and \underline{underline}, respectively. Methods in the upper block are trained and evaluated in the single-category setting, while methods in the lower block use a unified model for the multi-category setting across all 12 categories.}
\resizebox{\textwidth}{!}{
\begin{tabular}{l|cccccccccccc|c}
\toprule
\multicolumn{14}{c}{O-AUROC($\uparrow$)} \\ \midrule
Method & Airplane & Car & Candybar & Chicken & Diamond & Duck & Fish & Gemstone & Seahorse & Shell & Starfish & Toffees & Mean \\ 
\midrule
\multicolumn{14}{c}{\textit{Category-Specific Method}}\\
\midrule
BTF(Raw)             & 73.0 & 64.7 & 53.9 & 78.9 & 70.7 & 69.1 & 60.2 & \underline{68.6} & 59.6 & 39.6 & 53.0 & 70.3 & 63.5 \\
BTF(FPFH)            & 52.0 & 56.0 & 63.0 & 43.2 & 54.5 & 78.4 & 54.9 & 64.8 & \underline{77.9} & 75.4 & 57.5 & 46.2 & 60.3 \\
M3DM                 & 43.4 & 54.1 & 55.2 & 68.3 & 60.2 & 43.3 & 54.0 & 64.4 & 49.5 & 69.4 & 55.1 & 45.0 & 59.4 \\
PatchCore(FPFH)      & \textbf{88.2} & 59.0 & 54.1 & \underline{83.7} & 57.4 & 54.6 & 67.5 & 37.0 & 50.5 & 58.9 & 44.1 & 56.5 & 59.3 \\
PatchCore(PointMAE)  & 72.6 & 49.8 & 66.3 & 82.7 & 78.3 & 48.9 & 63.0 & 37.4 & 53.9 & 50.1 & 51.9 & 58.5 & 59.4 \\
CPMF                 & 70.1 & 55.1 & 55.2 & 50.4 & 52.3 & 58.2 & 55.8 & 58.9 & 72.9 & 65.3 & 70.0 & 39.0 & 58.6 \\
IMRNet               & 76.2 & 71.1 & 75.5 & 78.0 & 90.5 & 51.7 & 88.0 & 67.4 & 60.4 & 66.5 & 67.4 & 77.4 & 72.5 \\
Reg3D-AD             & 71.6 & 69.7 & 68.5 & \textbf{85.2} & 90.0 & 58.4 & 91.5 & 41.7 & 76.2 & 58.3 & 50.6 & 82.7 & 70.4 \\
Group3AD             & 74.4 & \underline{72.8} & \underline{84.7} & 78.6 & \underline{93.2} & 67.9 & \textbf{97.6} & 53.9 & \textbf{84.1} & 58.5 & 56.2 & \underline{79.6} & 75.1 \\
R3D-AD               & 77.2 & 69.6 & 71.3 & 71.4 & 68.5 & \textbf{90.9} & 69.2 & 66.5 & 72.0 & \textbf{84.0} & \underline{70.1} & 70.3 & 73.4 \\
ISMP                 & \underline{85.8} & \textbf{73.1} & \textbf{85.2} & 71.4 & \textbf{94.8} & 71.2 & \underline{94.5} & 46.8 & 72.9 & 62.3 & 66.0 & \textbf{84.2} & \textbf{76.7} \\
PO3AD                & 80.4 & 65.4 & 78.5 & 68.6 & 80.1 & \underline{82.0} & 85.9 & \textbf{69.3} & 75.6 & \underline{80.0} & \textbf{75.8} & 77.1 & \underline{76.5} \\
\midrule
\multicolumn{14}{c}{\textit{Unified Method}}\\
\midrule
MC3D-AD              & 85.0 & \underline{74.9} & \textbf{83.0} & \underline{71.5} & \textbf{95.5} & \underline{83.1} & \underline{86.5} & \underline{56.0} & \textbf{71.6} & \textbf{80.3} & \underline{76.6} & \underline{73.8} & \underline{78.2} \\
\rowcolor{gray!20}
\textbf{SeDiR (Ours)}        & \textbf{86.0} & \textbf{78.3} & \underline{81.9} & \textbf{72.9} & \underline{94.8} & \textbf{86.2} & \textbf{93.8} & \textbf{62.7} & \underline{67.4} & \underline{77.9} & \textbf{85.4} & \textbf{84.5} & \textbf{81.0} \\
\bottomrule
\end{tabular}
}
\label{tab:Real3D_combined}
\end{table*}
\renewcommand{\arraystretch}{1.08}

\subsection{Geometry-Guided Decoder}
\label{subsec:geom_decoder}

While CFGT and C3L establish semantically disentangled representations, 
the final stage of the proposed paradigm translates these category-aware priors into geometrically consistent reconstructions.
To this end, we introduce the Geometry-Guided Decoder (GGD), 
which reconstructs feature tokens conditioned on both the semantic prior and the input geometry.

Let $\mathbf{z}$ denote the global semantic prior and 
\(\tilde{\mathbf{F}}^{(k)}=\big[\tilde{\mathbf{f}}^{(k)}_{1},\ldots,\tilde{\mathbf{f}}^{(k)}_{g}\big]\) 
be the encoded $k$-resolution feature sequence. 
In GGD, $\mathbf{z}$ serves as the \emph{query} while \(\tilde{\mathbf{F}}^{(k)}\) provides the \emph{keys} and \emph{values}. 
To ensure semantically consistent reconstruction, we incorporate geometric cues directly into the attention mechanism by introducing a geometry-derived bias \(\mathbf{B}_{\mathrm{geo}}\), which encodes local geometric variations such as surface normal and curvature changes between point neighborhoods.
This bias is added to the attention logits to guide:
{\small\begin{equation}
\mathrm{Attention}(\mathbf{Q},\mathbf{K},\mathbf{V})
=\mathrm{softmax}\!\left(\frac{\mathbf{Q}\mathbf{K}^{\top}}{\sqrt{d}}+\beta\,\mathbf{B}_{\mathrm{geo}}\right)\mathbf{V},
\label{eq:geom_attn}
\end{equation}}
where \(\beta\) controls the strength of geometric guidance.
Details of $\mathbf{B}_{\mathrm{geo}}$ are provided in the supplementary.

As shown in \cref{fig:figure_4}, outputs from two such modules are concatenated and refined via a feed-forward network (FFN) 
to yield reconstructed feature tokens 
\(\hat{\mathbf{F}}^{(k)}=\big[\hat{\mathbf{f}}^{(k)}_{1},\ldots,\hat{\mathbf{f}}^{(k)}_{g}\big]\). 
Reconstruction is supervised by an MSE loss:
\begin{equation}
\mathcal{L}_{\mathrm{rec}}
=\frac{1}{g}\sum_{j=1}^{g}\lVert \hat{\mathbf{f}}^{(k)}_{j}-\mathbf{f}^{(k)}_{j}\rVert_2^{2}.
\end{equation}

\noindent\textbf{Final training objective.}
The overall objective combines semantic disentanglement and geometry-guided reconstruction:
\begin{equation}
\mathcal{L}_{\text{total}} =
\mathcal{L}_{\text{C3L}} + \mathcal{L}_{\text{rec}},
\end{equation}
where $\mathcal{L}_{\text{C3L}}$ (\cref{eq:C3L_loss}) enforces category-conditioned separation 
and $\mathcal{L}_{\text{rec}}$ ensures geometric fidelity. 
This joint optimization produces reconstructions that are both semantically consistent and geometrically accurate, 
fulfilling the goal of Semantically Disentangled Reconstruction.


\subsection{Anomaly Scoring}
\label{subsec:anomaly_score}

At inference, a test point cloud is encoded into feature tokens and reconstructed by the decoder.
To quantify the reconstruction discrepancy, we compute the $\ell_2$ distance between the original and reconstructed tokens and normalize it within each instance:
\begin{equation}
\mathbf{S}_p
=\operatorname{Gauss}_{k_g,\sigma}\!\Big(
\operatorname{Norm}(\lVert \hat{\mathbf{F}}^{(k)}-{\mathbf{F}}^{(k)}\rVert_2)
\Big),
\end{equation}
where $\operatorname{Norm}$ denotes min–max normalization and $\operatorname{Gauss}_{k_g,\sigma}$ is a discrete 1D Gaussian filter applied along the token dimension. The per-point scores \(\mathbf{S}_p\) indicate local inconsistencies, and the object-level anomaly score is defined as 
$S_{\mathrm{obj}}=\max(\mathbf{S}_p)$.
\section{Experiments}
\label{sec:exp}

\setlength{\abovecaptionskip}{3pt}
\setlength{\belowcaptionskip}{0pt}

\subsection{Experimental Settings}
\noindent\textbf{Datasets.}
We evaluate our approach on two benchmark datasets: Real3D-AD~\cite{liu2023real3d} and Anomaly-ShapeNet~\cite{li2024towards}. Anomaly-ShapeNet is a synthetic 3D anomaly detection dataset containing 1,600 samples across 40 categories, with 4 normal samples per category in the training set.
Real3D-AD is a high-resolution real-world dataset with 12 categories, each with 4 normal training samples and 100 test samples. While both datasets employ complete (360°) point clouds for training, Real3D-AD provides single-view test scans, making anomaly detection more challenging.

\noindent\textbf{Evaluation metrics.}
Following prior work~\cite{liu2023real3d, li2024towards}, we evaluate anomaly detection performance using the Area Under the Receiver Operating Characteristic Curve (AUROC). We report O-AUROC for object-level anomaly detection and P-AUROC for point-level anomaly localization. Higher O-AUROC and P-AUROC indicate better detection and localization performance, respectively.

\noindent\textbf{Baseline methods.}
We compare our method against state-of-the-art approaches, including BTF~\cite{horwitz2023back}, M3DM~\cite{wang2023multimodal}, PatchCore~\cite{roth2022towards}, CPMF~\cite{cao2024complementary}, IMRNet~\cite{li2024towards}, Reg3D-AD~\cite{liu2023real3d}, Group3AD~\cite{zhu2024towards}, R3D-AD~\cite{zhou2024r3d}, ISMP~\cite{liang2025look}, PO3AD~\cite{ye2025po3ad}, and MC3D-AD~\cite{ijcai2025p94}. MC3D-AD trains a unified model for all categories, while the remaining methods train individual models per category. The results of these methods are obtained from publicly available code or their papers.

\subsection{Comparison Results with SOTA Methods}

\noindent\textbf{Object-level results on Real3D-AD.}
As shown in \cref{tab:Real3D_combined}, our method achieves state-of-the-art performance on O-AUROC, outperforming the second-best approach by 2.8\% across both single- and multi-category settings. 
This consistent improvement indicates that our framework effectively learns category-disentangled features, leading to robust 3D anomaly detection across diverse category distributions.

\noindent\textbf{Object-level results on Anomaly-ShapeNet.}
\cref{tab:Anomaly_object} reports the object-level AUROC results on the Anomaly-ShapeNet dataset. 
Our method achieves 93.3\% O-AUROC, surpassing the second-best approach by 9.1\%. 
These results demonstrate that discriminative, category-aware representations substantially improve generalization in the unified multi-category setting, enabling more accurate and semantically consistent anomaly detection and localization.

\noindent\textbf{Point-level results.}
As reported in \cref{tab:pauroc_means_two_datasets}, our method achieves the second-best P-AUROC on both Real3D-AD and Anomaly-ShapeNet datasets, while single-category approaches exhibit dataset-specific strengths.
ISMP~\cite{liang2025look}, a memory-based method, performs best on Real3D-AD by effectively recalling detailed local patterns in real-world data, but its reliance on memorized normal features limits its generalization to synthetic domains (\eg, Anomaly-ShapeNet).
PO3AD~\cite{ye2025po3ad}, which generates pseudo-anomalies for supervision, achieves strong results on the synthetic Anomaly-ShapeNet dataset but struggles to generalize to real-world geometries in Real3D-AD.
In contrast, our framework maintains consistently high performance across both datasets, demonstrating superior robustness and generalization under the unified multi-category setting.

\begin{table*}[t!]
\caption{Comparison of object-level AUROC (\%) of various methods on the Anomaly-ShapeNet.}
\resizebox{\textwidth}{!}{
\begin{tabular}{l|cccccccccccccccccccc}
\toprule
\multicolumn{21}{c}{O-AUROC($\uparrow$)}\\
\midrule
Method & ashtray0 & bag0 & bottle0 & bottle1 & bottle3 & bowl0 & bowl1 & bowl2 & bowl3 & bowl4 & bowl5 & bucket0 & bucket1 & cap0 & cap3 & cap4 & cap5 & cup0 & cup1 & eraser0 \\
\midrule
\multicolumn{21}{c}{\textit{Category-Specific Method}}\\
\midrule
BTF(Raw)            & 57.8 & 41.0 & 59.7 & 51.0 & 56.8 & 56.4 & 26.4 & 52.5 & 38.5 & 66.4 & 41.7 & 61.7 & 32.1 & 66.8 & 52.7 & 46.8 & 37.3 & 40.3 & 52.1 & 52.5 \\
BTF(FPFH)           & 42.0 & 54.6 & 34.4 & 54.6 & 32.2 & 50.9 & 66.8 & 51.0 & 49.0 & 60.9 & 69.9 & 40.1 & 63.3 & 61.8 & 52.2 & 52.0 & 58.6 & 58.6 & 61.0 & 71.9 \\
M3DM                & 57.7 & 53.7 & 57.4 & 63.7 & 54.1 & 63.4 & 66.3 & 68.4 & 61.7 & 46.4 & 40.9 & 30.9 & 50.1 & 55.7 & 42.3 & 77.7 & 63.9 & 53.9 & 55.6 & 62.7 \\
PatchCore(FPFH)     & 58.7 & 57.1 & 60.4 & 66.7 & 57.2 & 50.4 & 63.9 & 61.5 & 53.7 & 49.4 & 55.8 & 46.9 & 55.1 & 58.0 & 45.3 & 75.7 & 79.0 & 60.0 & 58.6 & 65.7 \\
PatchCore(PointMAE) & 59.1 & 60.1 & 51.3 & 60.1 & 65.0 & 52.3 & 62.9 & 45.8 & 57.9 & 50.1 & 59.3 & 59.3 & 56.1 & 58.9 & 47.6 & 72.7 & 53.8 & 61.0 & 55.6 & 67.7 \\
CPMF                & 35.3 & 64.3 & 52.0 & 48.2 & 40.5 & 78.3 & 63.9 & 62.5 & 65.8 & 68.3 & 68.5 & 48.2 & 60.1 & 60.1 & 55.1 & 55.3 & \textbf{69.7} & 49.7 & 49.9 & 68.9 \\
Reg3D-AD            & 59.7 & 70.6 & 48.6 & 69.5 & 52.5 & 67.1 & 52.5 & 49.0 & 34.8 & 66.3 & 59.3 & 61.0 & 75.2 & 69.3 & 72.5 & 64.3 & 46.7 & 51.0 & 53.8 & 34.3 \\
IMRNet              & 67.1 & 66.0 & 55.2 & 70.0 & 64.0 & 68.1 & 70.2 & 68.5 & 59.9 & 67.6 & \underline{71.0} & 58.0 & \underline{77.1} & 73.7 & \underline{77.5} & 65.2 & 65.2 & 64.3 & \underline{75.7} & 54.8 \\
R3D-AD              & \underline{83.3} & \underline{72.0} & \underline{73.3} & \underline{73.7} & \underline{78.1} & \underline{81.9} & \underline{77.8} & \underline{74.1} & \underline{76.7} & \underline{74.4} & 65.6 & \underline{68.3} & 75.6 & \underline{82.2} & 73.0 & \underline{68.1} & \underline{67.0} & \underline{77.6} & \underline{75.7} & \underline{89.0} \\
PO3AD               & \textbf{100.0} & \textbf{83.3} & \textbf{90.0} & \textbf{93.3} & \textbf{92.6} & \textbf{92.2} & \textbf{82.9} & \textbf{83.3} & \textbf{88.1} & \textbf{98.1} & \textbf{84.9} & \textbf{85.3} & \textbf{78.7} & \textbf{87.7} & \textbf{85.9} & \textbf{79.2} & \underline{67.0} & \textbf{87.1} & \textbf{83.3} & \textbf{99.5} \\
\midrule
\multicolumn{21}{c}{\textit{Unified Method}}\\
\midrule
MC3D-AD             & \underline{96.2} & \underline{80.5} & \underline{79.5} & \underline{70.9} & \underline{75.6} & \underline{93.0} & \textbf{97.8} & \underline{71.9} & \underline{88.5} & \underline{91.1} & \underline{75.4} & \underline{89.8} & \underline{78.4} & \underline{79.3} & \underline{70.1} & \underline{83.5} & \underline{76.1} & \underline{74.3} & \underline{95.2} & \textbf{77.6} \\
\rowcolor{gray!20}
\textbf{SeDiR (Ours)}       & \textbf{97.6} & \textbf{88.6} & \textbf{92.9} & \textbf{91.6} & \textbf{98.4} & \textbf{95.9} & \underline{95.6} & \textbf{92.6} & \textbf{96.7} & \textbf{98.5} & \textbf{97.9} & \textbf{91.1} & \textbf{93.3} & \textbf{98.5} & \textbf{98.9} & \textbf{99.6} & \textbf{87.4} & \textbf{99.5} & \textbf{100.0} & \underline{66.7} \\
\bottomrule
\end{tabular}
}
\resizebox{\textwidth}{!}{
\begin{tabular}{l|cccccccccccccccccccc|c}
\toprule
Method & headset0 & headset1 & helmet0 & helmet1 & helmet2 & helmet3 & jar & phone & shelf0 & tap0 & tap1 & vase0 & vase1 & vase2 & vase3 & vase4 & vase5 & vase7 & vase8 & vase9 & Mean \\
\midrule
\multicolumn{21}{c}{\textit{Category-Specific Method}}\\
\midrule
BTF(Raw)            & 37.8 & 51.5 & 55.3 & 34.9 & 60.2 & 52.6 & 42.0 & 56.3 & 16.4 & 52.5 & 57.3 & 53.1 & 54.9 & 41.0 & 71.7 & 42.5 & 58.5 & 44.8 & 42.4 & 56.4 & 49.3 \\
BTF(FPFH)           & 52.0 & 49.0 & 57.1 & 71.9 & 54.2 & 44.4 & 58.6 & 67.1 & 60.9 & 56.0 & 54.6 & 34.2 & 21.9 & 54.6 & 69.9 & 51.0 & 40.9 & 51.8 & 66.8 & 26.8 & 52.8 \\
M3DM                & 57.7 & 61.7 & 52.6 & 42.7 & 62.3 & 37.4 & 56.4 & 35.7 & 56.4 & \textbf{75.4} & 73.9 & 57.4 & 42.7 & 73.7 & 43.9 & 47.6 & 43.9 & 65.7 & 66.3 & 66.3 & 55.2 \\
PatchCore(FPFH)     & 58.3 & 63.7 & 54.6 & 48.4 & 42.5 & 40.4 & 49.4 & 38.8 & 49.4 & \underline{75.3} & \underline{76.6} & 60.4 & 42.3 & 72.1 & 44.9 & 50.6 & 44.9 & 69.3 & 66.2 & 66.0 & 56.8 \\
PatchCore(PointMAE) & 59.1 & 62.7 & 55.6 & 55.2 & 44.7 & 42.4 & 48.3 & 48.8 & 52.3 & 45.8 & 53.8 & 51.3 & 55.2 & 74.1 & 46.0 & 51.6 & 46.0 & 65.0 & 66.3 & 62.9 & 56.2 \\
CPMF                & 64.3 & 45.8 & 55.5 & 58.9 & 46.2 & 52.0 & 61.0 & 50.9 & 68.5 & 35.9 & 69.7 & 58.2 & 34.5 & 58.2 & 58.2 & 51.4 & 58.2 & 39.7 & 52.9 & 60.9 & 55.9 \\
Reg3D-AD            & 53.7 & 61.0 & 60.0 & 38.1 & 61.4 & 36.7 & 59.2 & 41.4 & \underline{68.8} & 67.6 & 64.1 & 48.6 & 70.2 & 60.5 & 65.0 & 50.0 & 65.0 & 46.2 & 62.0 & 59.4 & 57.2 \\
IMRNet              & 72.0 & 67.6 & 59.7 & 60.0 & \underline{64.1} & 57.3 & 78.0 & 75.5 & 60.3 & 67.6 & 69.6 & 55.2 & \textbf{75.7} & 61.4 & 70.0 & 52.4 & 70.0 & 63.5 & 63.0 & 59.4 & 66.1 \\
R3D-AD              & \underline{73.8} & \underline{79.5} & \underline{75.7} & \underline{72.0} & 63.3 & \underline{70.7} & \underline{83.8} & \underline{76.2} & \textbf{69.6} & 73.6 & \textbf{90.0} & \underline{73.3} & 72.9 & \underline{75.2} & \underline{74.2} & \underline{63.0} & \underline{74.2} & \underline{77.1} & \underline{72.1} & \underline{71.8} & \underline{74.9} \\
PO3AD               & \textbf{80.8} & \textbf{92.3} & \textbf{76.2} & \textbf{96.1} & \textbf{86.9} & \textbf{75.4} & \textbf{86.6} & \textbf{77.6} & 57.3 & 74.5 & 68.1 & \textbf{85.8} & \underline{74.2} & \textbf{95.2} & \textbf{82.1} & \textbf{67.5} & \textbf{85.2} & \textbf{96.6} & \textbf{73.9} & \textbf{83.0} & \textbf{83.9} \\
\midrule
\multicolumn{21}{c}{\textit{Unified Method}}\\
\midrule
MC3D-AD             & \textbf{86.2} & \underline{88.6} & \underline{67.2} & \textbf{100.0} & \underline{60.9} & \underline{97.9} & \underline{97.1} & \textbf{91.9} & \underline{84.1} & \textbf{94.5} & \textbf{97.0} & \underline{79.5} & \underline{85.7} & \underline{92.9} & \underline{76.1} & \underline{87.6} & \underline{76.1} & \underline{93.8} & \underline{67.0} & \underline{73.6} & \underline{84.2} \\
\rowcolor{gray!20}
\textbf{SeDiR (Ours)}       & \underline{84.9} & \textbf{89.0} & \textbf{96.5} & \textbf{100.0} & \textbf{87.0} & \textbf{100.0} & \textbf{100.0} & \underline{90.5} & \textbf{88.7} & \underline{92.7} & \underline{89.3} & \textbf{88.8} & \textbf{97.1} & \textbf{97.6} & \textbf{78.8} & \textbf{92.4} & \textbf{94.8} & \textbf{100.0} & \textbf{87.3} & \textbf{94.2} & \textbf{93.3} \\
\bottomrule
\end{tabular}
}
\label{tab:Anomaly_object}
\vspace{-4mm}
\end{table*}

\begin{table}[t!]
  \centering
  \caption{
Comparison of mean point-level AUROC (\%) on Real3D-AD and Anomaly-ShapeNet datasets.
The detailed per-category results are provided in the supplementary.
}
\label{tab:pauroc_means_two_datasets}
\resizebox{0.8\linewidth}{!}{
\begin{tabular}{l|cc}
\toprule
Method & Real3D-AD & Anomaly-ShapeNet \\ 
\midrule
\multicolumn{3}{c}{\textit{Category-Specific Method}} \\
\midrule
BTF(Raw)             & 57.1 & 55.0 \\
BTF(FPFH)            & 73.3 & 62.8 \\
M3DM                 & 62.0 & 61.6 \\
PatchCore(FPFH)      & 68.2 & 58.0 \\
PatchCore(PointMAE)  & 62.0 & 57.7 \\
CPMF                 & \underline{75.8} & 57.3 \\
IMRNet               & --   & 65.0 \\
Reg3D-AD             & 70.5 & 66.8 \\
Group3AD             & 73.5 & -- \\
ISMP                 & \textbf{83.6} & \underline{69.1} \\
PO3AD                & 65.0 & \textbf{89.8} \\
\midrule
\multicolumn{3}{c}{\textit{Unified Method}} \\
\midrule
MC3D-AD              & \underline{76.8} & \underline{75.8} \\
\rowcolor{gray!10}
\textbf{SeDiR (Ours)}                 
& \textbf{80.6} & \textbf{81.0} \\
\bottomrule
\end{tabular}}
\vspace{-4mm}
\end{table}

\subsection{Ablation Studies}
\label{subsec:ablation}
We perform comprehensive ablation experiments to quantify the contribution of each proposed component to the performance of our model. These are shown across \cref{tab:ablation_entire} through \cref{tab:ablation_gt}.

\noindent\textbf{Effect of each component.}
\cref{tab:ablation_entire} demonstrates the impact of the three proposed modules: CFGT, C3L, and GGD.
Incorporating CFGT yields a slight performance gain, as fusing multi-resolution geometric cues with an adaptive context token helps form a category-aware global representation.
Adding C3L improves O-AUROC by 2–3\%, as it reduces inter-category entanglement and yields more discriminative embeddings.
Finally, GGD brings another 1–2\%, showing that geometry-guided reconstruction encourages semantically consistent outputs.
These results collectively validate that each component contributes complementarily and that the overall framework effectively realizes the idea of Semantically Disentangled Reconstruction.

\begin{table}[t]
\centering
\caption{Ablation studies on the effectiveness of core components.}
\resizebox{0.7\linewidth}{!}
{
\begin{tabular}{@{}ccccc@{}}
\toprule
CFGT & C3L & GGD & \makecell{O-AUROC ($\uparrow$)} & \makecell{P-AUROC ($\uparrow$)}  \\
\midrule
& & & 75.1 & 78.1 \\
\checkmark & & & 76.4 & 78.4 \\
\checkmark & \checkmark & & 79.2 & 79.0 \\
\checkmark & \checkmark & \checkmark & \textbf{81.0} & \textbf{80.6} \\
\bottomrule
\end{tabular}
}
\label{tab:ablation_entire}
\vspace{-2mm}
\end{table}

\begin{figure*}[!t] 
    \centerline{\includegraphics[width=\linewidth]{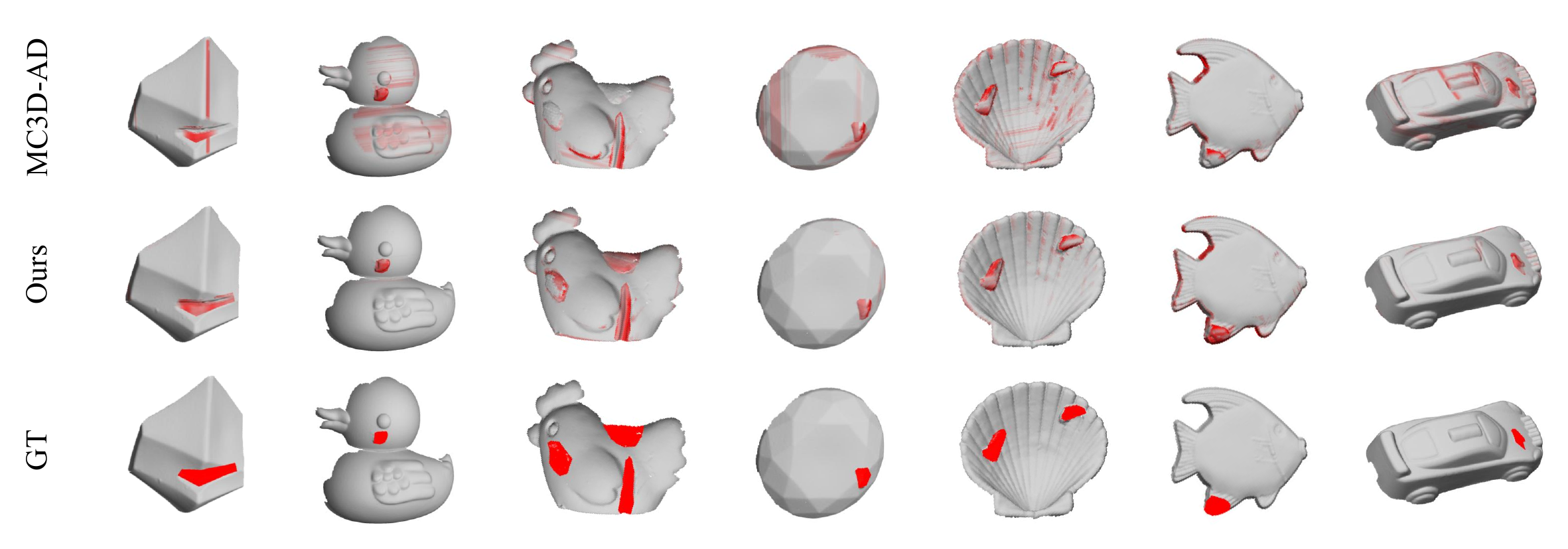}}
    \caption{Qualitative comparison with MC3D-AD across various object categories in Real3D-AD dataset. Red colored regions indicate abnormalities (\eg, bulge, sink). MC3D-AD often misses true anomalies or yields false positives, whereas our method delivers more accurate and complete localization aligned with the ground truth.}
    \label{fig:figure_5} 
    \vspace{-4mm}
\end{figure*}

\begin{table}[h]
\centering
\caption{Ablation studies on the contribution of each loss in C3L.}
\resizebox{0.7\linewidth}{!}
{
\begin{tabular}{@{}cccccc@{}}
\toprule
$\mathcal{L}_{\text{scl}}$ & $\mathcal{L}_{\text{cls}}$ & $\mathcal{L}_{\text{cos}}$ & \makecell{O-AUROC ($\uparrow$)} & \makecell{P-AUROC ($\uparrow$)}  \\
\midrule
\checkmark & & & 79.1 & 78.8 \\
\checkmark & \checkmark & & 80.0 & 79.3 \\
\checkmark & \checkmark &  \checkmark & \textbf{81.0} & \textbf{80.6} \\
\bottomrule
\end{tabular}
}
\label{tab:ablation_c3l}
\vspace{-4mm}
\end{table}

\noindent\textbf{Analysis of loss components in C3L.}
\cref{tab:ablation_c3l} analyzes the effects of the three loss components in the C3L module—namely, the supervised contrastive loss ($\mathcal{L}_{\text{scl}}$), the auxiliary classification loss ($\mathcal{L}_{\text{cls}}$), and the multi-resolution cosine alignment loss ($\mathcal{L}_{\text{cos}}$).
Using only $\mathcal{L}_{\text{scl}}$ achieves reasonable performance, showing that contrastive learning alone provides meaningful feature separation.
Adding $\mathcal{L}_{\text{cls}}$ yields a slight improvement, confirming that explicit category supervision enhances semantic discrimination across object classes.
Finally, adding $\mathcal{L}_{\text{cos}}$ yields a further 1\% gain, showing that cross-resolution consistency enhances representation stability.
Overall, combining the three losses reinforces semantic disentanglement and preserves intra-category coherence, maximizing detection performance.

\begin{table}[h]
\footnotesize
\centering
\caption{Comparison of different guidance strategies in GGD.}
\begin{tabular}{@{}lcc@{}}
\toprule
{Method}               & {O-AUROC} ($\uparrow$) & {P-AUROC} ($\uparrow$) \\ \midrule
Masking             & 79.8 & 79.6   \\
Gating     & 79.7 & 78.9   \\
\textbf{Bias (Ours)}            & \textbf{81.0} & \textbf{80.6}  \\ \bottomrule
\end{tabular}
\label{tab:ablation_geo}
\vspace{-4mm}
\end{table}

\noindent\textbf{Geometry-Guided strategies in GGD.}
\cref{tab:ablation_geo} compares different strategies for injecting geometric priors within the Geometry-Guided Decoder (GGD).
Our bias-based design achieves the best performance, while masking (as in MC3D-AD) or gating geometric priors leads to lower performance.
This indicates that incorporating geometric cues as an attention bias provides a more stable and fine-grained integration of spatial structure.
This bias formulation more effectively enforces geometry-consistent reconstruction under category semantics, resulting in improved anomaly detection and localization.

\begin{table}[h]
\footnotesize
\centering
\caption{Comparison of different global token strategies.}
\begin{tabular}{@{}lcc@{}}
\toprule
{Method}               & {O-AUROC} ($\uparrow$) & {P-AUROC} ($\uparrow$) \\ \midrule
ACT             & 78.5 & 78.5   \\
Max     &  76.4 & 75.5  \\
Mean     & 78.2 & 78.7   \\
\textbf{CFGT (Ours)}           & \textbf{81.0} & \textbf{80.6}   \\ \bottomrule
\end{tabular}
\label{tab:ablation_gt}
\vspace{-4mm}
\end{table}

\noindent\textbf{Comparison of global token configuration strategies.}
\cref{tab:ablation_gt} compares alternative strategies for constructing the global token in CFGT.
Replacing the fusion formulation in \cref{eq:global_concat} with either the ACT alone or simple pooling operations (Max or Mean) leads to clear performance drops, showing that these variants fail to capture sufficient contextual and spatial information.
In contrast, the proposed CFGT achieves the highest O-AUROC and P-AUROC by integrating multi-resolution geometric features with the adaptive context, producing a richer and more discriminative global representation.
These results demonstrate that CFGT provides stronger semantic consistency and discriminability than conventional global summarization methods.

\subsection{Qualitative Results}
\cref{fig:figure_5} shows qualitative comparisons between MC3D-AD and our method across various object categories in  Real3D-AD.
MC3D-AD often fails to accurately highlight the true anomalous regions, frequently missing some defective areas or producing false activations on normal surfaces.
In contrast, the proposed method provides more precise localization of anomalies.
This improvement suggests that incorporating semantic identity–aware representation learning and semantically guided reconstruction allows the model to better capture category-consistent semantics, leading to more reliable anomaly localization across diverse objects.
Additional qualitative results are provided in the supplementary.
\vspace{-6mm}
\section{Conclusion} 
\label{sec:conclusion}
\vspace{-1mm}
In this paper, we presented a unified framework for multi-category 3D anomaly detection, termed Semantically Disentangled Reconstruction (SeDiR). Unlike conventional unified models that reconstruct before understanding what is being reconstructed, our method first establishes a clear semantic identity and then performs reconstruction. By combining Coarse-to-Fine Global Tokenization, Category-Conditioned Contrastive Learning, and the Geometry-Guided Decoder, the method learns disentangled, category-aware representations and produces semantically aligned, geometrically consistent reconstructions. Extensive experiments show that each component provides complementary gains—CFGT improves global semantic awareness, C3L enhances category separation, and GGD enforces geometry-consistent reconstruction—collectively achieving state-of-the-art results across multiple benchmarks. We believe these findings provide new insights into integrating semantic and geometric reasoning for robust 3D perception and open promising directions for cross-domain and open-set 3D anomaly detection.

\section*{Acknowledgments}
This work was supported by the National Research Foundation of Korea (NRF) grant funded by the Korea government(MSIT)(RS-2024-00456589) and Korea Planning \& Evaluation Institute of Industrial Technology(KEIT) grant funded by the Korea government(MOTIE) (No. RS2024-00442120, Development of AI technology capable of robustly recognizing abnormal and dangerous situations and behaviors during night and bad weather conditions).
{
    \small
    \bibliographystyle{ieeenat_fullname}
    \bibliography{main}
}

\clearpage
\maketitlesupplementary

\appendix
\section*{Contents}

\setcounter{table}{0}
\renewcommand{\thetable}{S\arabic{table}}

\setcounter{figure}{0}
\renewcommand{\thefigure}{S\arabic{figure}}

~\ref{sec:A}. Implementation Details

\noindent~\ref{sec:B}. Details of Geometry-Derived Bias

\noindent~\ref{sec:C}. More Quantitative Results

\noindent~\ref{sec:D}. More Ablation Studies

\noindent~\ref{sec:E}. More Qualitative Results

\noindent~\ref{sec:F}. Limitation

\section{Implementation Details}\label{sec:A}
\noindent\textbf{Model setup.}
We adopt PointMAE~\cite{pang2023masked}, pretrained on ModelNet40 (8k), as the local feature extractor $E(\cdot)$. Each point cloud is partitioned into $1024$ groups via FPS~\cite{qi2017pointnet++}, and multi-resolution neighborhoods are formed with sizes $\{128, 256, 512\}$. Positional embeddings follow Point Transformer~\cite{zhao2021point}, where $(x,y,z)$ coordinates are encoded by a two-layer MLP with ReLU. The shared transformer encoder in CFGT and the Geometry-Guided Decoder each contain 4 layers with 8 attention heads. Before entering the transformer encoder, feature jittering is applied with scale 20.0 and probability 1.0. The buffer in C3L is set to 64 entries.

\noindent\textbf{Training and inference setup.}
We train the model using AdamW~\cite{loshchilov2017decoupled} with a base learning rate of $1\times10^{-4}$ and a cosine annealing schedule. The batch size is 1 and training is performed for 1000 epochs. The loss weights in C3L are set to $\lambda_{\mathrm{scl}} = 0.001$, $\lambda_{\mathrm{cls}} = 0.001$, and $\lambda_{\mathrm{cos}} = 0.01$. 
During inference, reconstruction discrepancies are smoothed with a 1D Gaussian filter with kernel size $k_g = 511$ and standard deviation $\sigma = 0.2$. All experiments are conducted with PyTorch 1.13.0 and CUDA 11.7 on a single NVIDIA GeForce RTX 3090 GPU.


\section{Details of Geometry-Derived Bias}\label{sec:B}

Given a point cloud $\mathbf{P}=\{\mathbf{p}_n\}_{n=1}^{N}$, 
Farthest Point Sampling~\cite{qi2017pointnet++} selects $g$ group centers 
$\mathcal{S}=\{\mathbf{s}_i\}_{i=1}^{g}$.  
For each center $\mathbf{s}_i$, an adaptive radius $r$ is estimated per instance, and its neighborhood is defined as
$\mathcal{N}_i = \{\mathbf{p}_j \mid \lVert \mathbf{p}_j - \mathbf{s}_i \rVert \le r\}$.

\noindent\textbf{Normal and curvature estimation.}
Let $\mathbf{C}_i$ denote the covariance matrix of centered neighbors:
\begin{equation}
\mathbf{C}_i = \mathrm{Cov}\!\big(\mathcal{N}_i - \operatorname{mean}(\mathcal{N}_i)\big).
\end{equation}
Eigen-decomposition produces eigenvalues $\lambda_0 \le \lambda_1 \le \lambda_2$ and associated eigenvectors $\mathbf{e}_0,\mathbf{e}_1,\mathbf{e}_2$.  
The surface normal and curvature at the group center $\mathbf{s}_i$ are obtained as:
\begin{equation}
\mathbf{n}_i = \mathbf{e}_0,\qquad
\kappa_i = \frac{\lambda_0}{\lambda_0+\lambda_1+\lambda_2}.
\end{equation}

\noindent\textbf{Local geometric variations.}
Geometric irregularity is quantified through angular deviations of normals and curvature fluctuations within each neighborhood:
\begin{equation}
v_i^{\mathrm{norm}}
= \frac{1}{|\mathcal{N}_i|}
\sum_{\mathbf{p}_j \in \mathcal{N}_i}
\angle(\mathbf{n}_i,\mathbf{n}_j),
\end{equation}
\begin{equation}
v_i^{\mathrm{curv}}
= \frac{1}{|\mathcal{N}_i|}
\sum_{\mathbf{p}_j \in \mathcal{N}_i}
|\kappa_i - \kappa_j|.
\end{equation}
These two measures form a geometric descriptor:
\begin{equation}
\mathbf{v}_i = [v_i^{\mathrm{norm}},\, v_i^{\mathrm{curv}}].
\end{equation}
After normalization, the descriptor is passed through a lightweight MLP to produce a scalar bias:
\begin{equation}
\mathbf{B}_{\mathrm{geo}}(i) = \mathrm{MLP}({\mathbf{v}}_i).
\end{equation}
The lightweight MLP is a two-layer MLP with 
ReLU.

\section{More Quantitative Results}\label{sec:C}
\noindent\textbf{P-AUROC results.}
\cref{tab:real3d_category} and \cref{tab:asnet_category} present the full category-level results for Real3D-AD~\cite{liu2023real3d} and Anomaly-ShapeNet~\cite{li2024towards}, respectively.
While PO3AD~\cite{ye2025po3ad} and ISMP~\cite{liang2025look} achieve strong performance on specific datasets—excelling on synthetic and real-world domains respectively—our method maintains consistently solid performance across both datasets under the unified multi-category setting.
These results highlight the robustness and general applicability of our framework across diverse object categories and domains.

\noindent\textbf{O-AUPR results on Anomaly-ShapeNet.}
We additionally report object-level AUPR on the Anomaly-ShapeNet dataset in \cref{tab:o_aupr_as}.
Our method achieves strong performance across categories, outperforming existing approaches.
These results further confirm that the proposed semantically disentangled reconstruction contributes to reliable object-level anomaly detection under the unified multi-category setting.

\begin{table}[h]
\footnotesize
\centering
\caption{Comparison of computational cost.}
\begin{tabular}{@{}lcc@{}}
\toprule
Method              & Parameters (M) & FLOPs (G) \\ \midrule
MC3D-AD             & 34.16 & 439.21   \\
\textbf{Ours}            & 33.25 & 508.25   \\
\bottomrule
\end{tabular}
\label{tab:computational}
\end{table}

\noindent\textbf{Computational cost.}
\cref{tab:computational} compares the computational efficiency of our method with MC3D-AD~\cite{ijcai2025p94}.
Our model uses fewer parameters (33.25M vs.\ 34.16M), indicating a slightly lighter architecture.
Although the FLOPs are higher (508.25 vs.\ 439.21), this overhead mainly stems from the additional global tokenization and geometric bias modules.
Overall, the model achieves improved performance while maintaining comparable computational complexity.

\section{More Ablation Studies}\label{sec:D}
To better understand how each design choice contributes to our framework, we conduct extensive ablation studies across several key components on the Real3D-AD dataset.
These experiments provide a comprehensive analysis of how each factor influences semantic disentanglement, reconstruction quality, and overall anomaly detection performance.

\begin{table}[h]
\centering
\caption{Ablation studies on the CFGT and C3L losses.}
\resizebox{0.85\linewidth}{!}
{
\begin{tabular}{@{}c|ccc|cc@{}}
\toprule
CFGT & $\mathcal{L}_{\text{scl}}$ & $\mathcal{L}_{\text{cos}}$ & $\mathcal{L}_{\text{cls}}$ & \makecell{O-AUROC ($\uparrow$)} & \makecell{P-AUROC ($\uparrow$)}  \\
\midrule
& \checkmark & & & 78.8 & 79.0 \\
& & \checkmark & & 77.9 & 78.7 \\
& \checkmark & \checkmark &  & 80.5 & 78.6 \\
\midrule
\checkmark & \checkmark & & & 80.0 & 79.1 \\
\checkmark & & \checkmark & & 79.1 & 78.8 \\
\checkmark & & & \checkmark & 79.3 & 79.6 \\
\checkmark & \checkmark & \checkmark &  & 80.5 & 79.8 \\
\checkmark &  & \checkmark & \checkmark & 80.1 & 79.9 \\
\checkmark & \checkmark & & \checkmark  & 80.0 & 79.3 \\
\checkmark & \checkmark & \checkmark & \checkmark  & \textbf{81.0} & \textbf{80.6} \\
\bottomrule
\end{tabular}
}
\label{tab:ablation_c3l_cfgt}
\end{table}

\noindent\textbf{Ablation on the CFGT and C3L losses.}
\cref{tab:ablation_c3l_cfgt} reports the effect of enabling Coarse-to-Fine Global Tokenization (CFGT) module and the individual loss terms in C3L.
Without CFGT, no global token is formed; the model instead applies supervised contrastive learning to a globally averaged feature, resulting in reasonable yet suboptimal performance.
Enabling CFGT introduces a unified global token derived from multi-resolution features, which substantially strengthens category-conditioned contrastive learning.
Adding the cosine alignment and auxiliary classification terms further improves performance, with the full configuration achieving the highest O-AUROC and P-AUROC.
These results suggest that explicitly constructing a global token and supervising it with all C3L objectives yields the most discriminative and semantically disentangled representation.

\begin{table}[h]
\footnotesize
\centering
\caption{Comparison of different coarse-to-fine neighborhood ratios.}
\begin{tabular}{@{}lcc@{}}
\toprule
{Ratios}               & {O-AUROC} ($\uparrow$) & {P-AUROC} ($\uparrow$) \\ \midrule
\{k, 4k, k/4\}            & 80.3 & 79.4   \\
\{k, 3k, k/3\}             & 80.9 & 79.8 \\
\{k, 1.5k, 2k/3\}          & 79.9 & 79.8   \\
\textbf{\{k, 2k, k/2\}}    & \textbf{81.0} & \textbf{80.6}  \\ \bottomrule
\end{tabular}
\label{tab:CFGT_rate}
\end{table}

\noindent\textbf{Impact of coarse–fine neighborhood ratios.}
\cref{tab:CFGT_rate} reports the performance of different neighborhood ratio configurations used in the CFGT. 
We evaluate several coarse--fine combinations based on the base resolution $k$, including 
$\{k, 4k, k/4\}$, 
$\{k, 3k, k/3\}$, 
$\{k, 1.5k, 2k/3\}$, 
and $\{k, 2k, k/2\}$. 
Among these settings, the configuration $\{k, 2k, k/2\}$ achieves the best results on both O-AUROC and P-AUROC.
Across the evaluated ratios, settings with a more balanced fine--base--coarse distribution---such as $\{k, 2k, k/2\}$---tend to show more stable performance. 
This observation indicates that the choice of multi-resolution neighborhood arrangement can influence the effectiveness of global representation learning within CFGT.

\begin{table}[h]
\footnotesize
\centering
\caption{Comparison of different base neighborhood sizes $k$.}
\begin{tabular}{@{}lcc@{}}
\toprule
{Size}               & {O-AUROC} ($\uparrow$) & {P-AUROC} ($\uparrow$) \\ \midrule
\{256, \underline{512}, 1024\}            & 77.5 & 78.5   \\
\{64, \underline{128}, 256\}             & 79.9 & 79.6   \\
\textbf{\{128, \underline{256}, 512\}}    & \textbf{81.0} & \textbf{80.6}  \\ \bottomrule
\end{tabular}
\label{tab:neighborhood_size}
\end{table}

\noindent\textbf{Effect of neighborhood sizes.}
\cref{tab:neighborhood_size} evaluates different neighborhood sizes used in the CFGT module.
Three configurations with various \textit{base} sizes are compared, among which the setting with a \underline{base size} of 256 (\{128, \underline{256}, 512\}) achieves the highest O-AUROC and P-AUROC.
The results show that neighborhood size introduces performance gap: larger neighborhoods tend to smooth out fine geometric cues, whereas smaller ones may lack sufficient structural context.
A base size of 256 provides the most balanced fine–base–coarse composition, enabling stable multi-resolution feature extraction and yielding the most reliable performance among the evaluated choices.

\begin{table}[h]
\footnotesize
\centering
\caption{Comparison of different numbers of groups.}
\begin{tabular}{@{}lcc@{}}
\toprule
{N}               & {O-AUROC} ($\uparrow$) & {P-AUROC} ($\uparrow$) \\ \midrule
512            & 75.7 & 77.4   \\
2048             & 77.7 & 78.4   \\
\textbf{1024}    & \textbf{81.0} & \textbf{80.6}  \\ \bottomrule
\end{tabular}
\label{tab:num_groups}
\end{table}

\noindent\textbf{Effect of the number of groups.}
\cref{tab:num_groups} evaluates the impact of varying the number of groups used to partition the point cloud.
Among the three tested settings—512, 1024, and 2048 groups—the configuration with 1024 groups yields the best performance.
A clear performance gap emerges due to the trade-off in spatial granularity.
Using too few groups (e.g., 512) produces large spatial regions, causing each token to aggregate overly coarse geometry and reducing sensitivity to fine anomalies.
In contrast, using too many groups (e.g., 2048) results in highly fragmented neighborhoods that capture limited local structure and introduce noise, weakening multi-resolution feature extraction.
The configuration with 1024 groups achieves a balanced decomposition, providing sufficient geometric detail while maintaining stable local neighborhoods.
This result highlights the importance of selecting an appropriate grouping resolution for robust multi-resolution representation learning.

\begin{table}[h]
\footnotesize
\centering
\caption{Comparison of different buffer sizes in C3L.}
\begin{tabular}{@{}lcc@{}}
\toprule
{L}               & {O-AUROC} ($\uparrow$) & {P-AUROC} ($\uparrow$) \\ \midrule
32            & 79.6 & 79.8   \\
128             & 80.3 & 80.3   \\
10,000             & 80.5 & 79.2   \\
\textbf{64}    & \textbf{81.0} & \textbf{80.6}  \\ \bottomrule
\end{tabular}
\label{tab:c3l_buffer}
\end{table}

\noindent\textbf{Effect of buffer size in C3L.}
\cref{tab:c3l_buffer} examines the influence of buffer size in C3L.
Reducing the buffer from 64 to 32 slightly degrades both O-AUROC and P-AUROC, implying that too few contrastive samples hinder stable supervised contrastive learning.
Increasing the buffer (e.g., to 128 or 10{,}000) yields performance similar to that of size 64, suggesting diminishing returns once a moderate number of samples is available; excessively large buffers may even introduce stale embeddings that are misaligned with the current representation space.
Overall, a buffer size of 64 provides the best balance between stability and efficiency, yielding the highest performance among the evaluated configurations.

\begin{table}[h]
\centering
\caption{Comparison of different loss weights in C3L.}
\resizebox{0.7\linewidth}{!}
{
\begin{tabular}{@{}cccccc@{}}
\toprule
$\lambda_\mathrm{scl}$ & $\lambda_\mathrm{cos}$ & $\lambda_\mathrm{cls}$ & \makecell{O-AUROC ($\uparrow$)} & \makecell{P-AUROC ($\uparrow$)}  \\
\midrule
0.01 & 0.001 & 0.01 & 79.3 & 79.6 \\
0.001 & 0.01 & 0.01 & 78.6 & 78.3 \\
0.001 & 0.001 & 0.1 & 77.0 & 78.9 \\
\textbf{0.001} & \textbf{0.001} & \textbf{0.01} & \textbf{81.0} & \textbf{80.6} \\
\bottomrule
\end{tabular}
}
\label{tab:loss_abl}
\end{table}

\noindent\textbf{Effect of loss weights in C3L.}
\cref{tab:loss_abl} examines the sensitivity of C3L to different loss-weight configurations, obtained by scaling each term by a factor of 10.
Overweighting any single component disrupts the balance among the supervised contrastive, cosine alignment, and auxiliary classification losses, leading to noticeable drops in both O-AUROC and P-AUROC.
The weights used in the main experiments ($\lambda_{\mathrm{scl}}{=}0.001$, $\lambda_{\mathrm{cos}}{=}0.001$, $\lambda_{\mathrm{cls}}{=}0.01$) achieve the best performance, indicating that C3L benefits from a well-balanced combination of objectives.

\begin{table}[h]
\footnotesize
\centering
\caption{Category prediction accuracy with different global representations.}
\begin{tabular}{@{}lcc@{}}
\toprule
{Method}               & Accuracy (\%) \\ \midrule
Mean            & 78.1  \\
Max            & 77.3  \\
\textbf{ACT}           & \textbf{97.3}  \\ \bottomrule
\end{tabular}
\label{tab:cls_act}
\end{table}

\noindent\textbf{Effect of ACT for category representation.}
We evaluate the effectiveness of the proposed Adaptive Context Token (ACT) as a global summary of category information. After freezing the pretrained model, we train an additional category prediction head for 100 epochs using three different global representations: mean pooling, max pooling, and ACT. As shown in \cref{tab:cls_act}, ACT achieves substantially higher classification accuracy, indicating that it captures category-discriminative semantics far more effectively than conventional pooling-based global tokens. This validates the use of ACT as the primary representation for category-aware learning in our framework.


\section{More Qualitative Results}\label{sec:E}

\noindent\textbf{Qualitative comparison with MC3D-AD on Anomaly-ShapeNet.}
\cref{fig:supple_fig2} presents additional qualitative comparisons with MC3D-AD on the Anomaly-ShapeNet dataset. While MC3D-AD often yields incomplete or noisy anomaly responses, our method more distinctly highlights anomaly regions and produces more coherent anomaly maps. These results further verify the stability and effectiveness of our method across diverse geometric patterns.

\begin{figure}[h] 
    \centerline{\includegraphics[width=\linewidth]{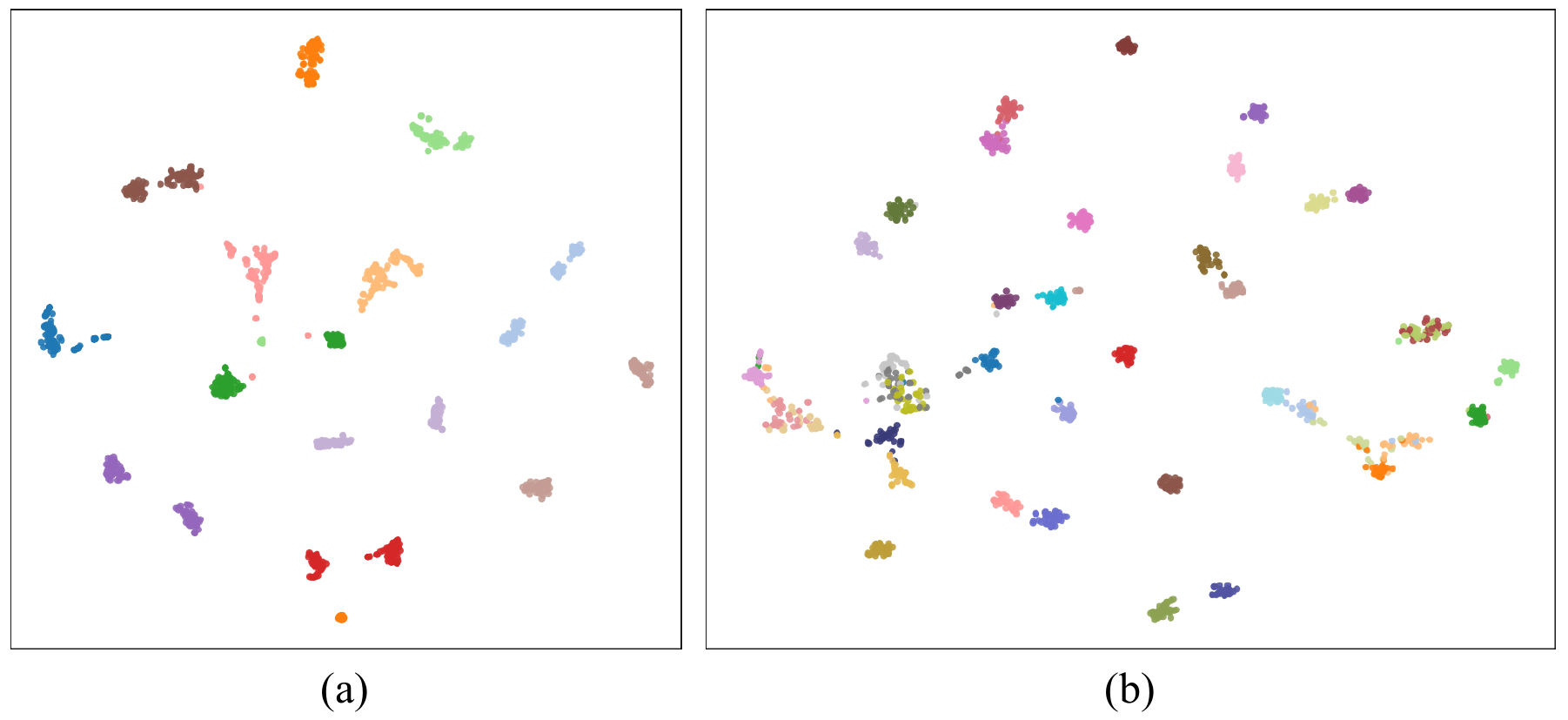}}
    \caption{t-SNE visualization of the learned global tokens on
(a) Real3D-AD (12 categories) and (b) Anomaly-ShapeNet (40 categories).
The embeddings form well-structured clusters, indicating that the proposed method learns coherent and category-aware global representations across both datasets.}
    \label{fig:sup_fig1} 
\end{figure}

\noindent\textbf{t-SNE visualization of global tokens.}
\Cref{fig:sup_fig1} presents the t-SNE~\cite{maaten2008visualizing} embeddings of the global tokens learned on Real3D-AD and Anomaly-ShapeNet. 
Across both datasets, the embeddings exhibit well-formed clusters that reflect meaningful category structure, demonstrating that CFGT and C3L effectively promote category-aware global representations. 
The consistent clustering pattern, even as the number of categories increases, highlights the robustness of the proposed global tokenization strategy.

\section{Limitation}\label{sec:F}
While our model adopts a category-aware design, it does not outperform MC3D-AD on every category. This behavior is expected, as our framework is built for class-aware generalization rather than class-specific tuning. Instead of learning highly specialized representations for individual categories, the model emphasizes semantic stability across categories, which inherently introduces trade-offs that may reduce performance in certain cases.
Nevertheless, this design choice provides clear advantages in overall robustness and generalization. Analyzing the per-category AUROC variance reveals that MC3D-AD exhibits considerably larger fluctuations, whereas our method yields much lower variance and thus more uniform performance. On the Real3D-AD dataset, the O-AUROC variances of MC3D-AD and ours are 90.0 and 86.7, respectively; on the Anomaly-ShapeNet dataset, MC3D-AD reaches 102.9, while ours remains substantially lower at 43.4. These observations indicate that our approach is less sensitive to category-specific biases and achieves more stable reconstruction in the multi-category setting.

\begin{table*}[!ht]
\caption{Comparison of point-level AUROC (\%) of various methods on the Real3D-AD dataset.}
\resizebox{\textwidth}{!}{
\begin{tabular}{l|cccccccccccc|c}
\toprule
\multicolumn{14}{c}{P-AUROC($\uparrow$)} \\ \midrule
Method & Airplane & Car & Candybar & Chicken & Diamond & Duck & Fish & Gemstone & Seahorse & Shell & Starfish & Toffees & Mean \\ 
\midrule
\multicolumn{14}{c}{\textit{Category-Specific Method}}\\
\midrule
   BTF(Raw)           & 56.4 & 64.7 & 73.5 & 60.9 & 56.3 & 60.1 & 51.4 & 59.7 & 52.0 & 48.9 & 39.2 & 62.3 & 57.1 \\
   BTF(FPFH)          & \underline{73.8} & 70.8 & \underline{86.4} & 73.5 & \underline{88.2} & \underline{87.5} & 70.9 & \underline{89.1} & 51.2 & 57.1 & 50.1 & 81.5 & 73.3 \\
   M3DM               & 54.7 & 60.2 & 67.9 & 67.8 & 60.8 & 66.7 & 60.6 & 67.4 & 56.0 & 73.8 & 53.2 & 68.2 & 62.0 \\
   PatchCore(FPFH)    & 56.2 & \underline{75.4} & 78.0 & 42.9 & 82.8 & 26.4 & 82.9 & \textbf{91.0} & 73.9 & 73.9 & 60.6 & 74.7 & 68.2 \\
   PatchCore(PointMAE)& 56.9 & 60.9 & 62.7 & 72.9 & 71.8 & 52.8 & 71.7 & 44.4 & 63.3 & 70.9 & 58.0 & 58.0 & 62.0 \\
   CPMF               & 61.8 & \textbf{83.6} & 73.4 & 55.9 & 75.3 & 71.9 & \textbf{98.8} & 44.9 & \textbf{96.2} & 72.5 & \textbf{80.0} & \textbf{95.9} & \underline{75.8} \\
   Reg3D-AD           & 63.1 & 71.8 & 72.4 & 67.6 & 83.5 & 50.3 & 82.6 & 54.5 & 81.7 & \underline{81.1} & 61.7 & 75.9 & 70.5 \\
   Group3AD           & 63.6 & 74.5 & 73.8 & \underline{75.9} & 86.2 & 63.1 & 83.6 & 56.4 & \underline{82.7} & 79.8 & 62.5 & 80.3 & 73.5 \\
   ISMP               & \textbf{75.3} & \textbf{83.6} & \textbf{90.7} & \textbf{79.8} & \textbf{92.6} & \textbf{87.6} & \underline{88.6} & 85.7 & 81.3 & \textbf{83.9} & 64.1 & \underline{89.5} & \textbf{83.6} \\
   PO3AD              & 71.5 & 57.4 & 76.3 & 56.3 & 61.4 & 64.4 & 87.7 & 53.0 & 58.5 & 54.2 & \underline{65.9} & 72.9 & 65.0 \\
\midrule
\multicolumn{14}{c}{\textit{Unified Method}}\\
\midrule
MC3D-AD            & \underline{62.8} & \underline{81.9} & \underline{91.0} & \textbf{64.0} & \underline{94.2} & \underline{82.2} & \underline{93.2} & \underline{45.8} & \underline{65.9} & \underline{77.8} & \underline{69.0} & \underline{93.4} & \underline{76.8} \\
\rowcolor{gray!20}
\textbf{Ours}      & \textbf{66.6} & \textbf{90.8} & \textbf{95.1} & \underline{63.2} & \textbf{97.7} & \textbf{83.4} & \textbf{96.7} & \textbf{54.7} & \textbf{73.9} & \textbf{79.8} & \textbf{70.9} & \textbf{94.0} & \textbf{80.6} \\
\bottomrule
\end{tabular}
}
\label{tab:real3d_category}
\end{table*}


\begin{table*}[t!]
\caption{Comparison of point-level AUROC (\%) of various methods on the Anomaly-ShapeNet dataset.}
\resizebox{\textwidth}{!}{
\begin{tabular}{l|ccccccccccccc}
\toprule
\multicolumn{14}{c}{P-AUROC($\uparrow$)}\\
\midrule
Method & ashtray0 & bag0 & bottle0 & bottle1 & bottle3 & bowl0 & bowl1 & bowl2 & bowl3 & bowl4 & bowl5 & bucket0 & bucket1 \\
\midrule
\multicolumn{14}{c}{\textit{Category-Specific Method}}\\
\midrule
BTF(Raw)            & 51.2 & 43.0 & 55.1 & 49.1 & \underline{72.0} & 52.4 & 46.4 & 42.6 & \underline{68.5} & 56.3 & 51.7 & 61.7 & 68.6 \\
BTF(FPFH)       & 62.4           & \underline{74.6}           & 64.1           & 54.9           & 62.2           & 71.0           & \underline{76.8}           & 51.8           & 59.0           & 67.9           & 69.9    & 40.1           & 63.3           \\
M3DM            & 57.7           & 63.7           & 66.3           & 63.7           & 53.2           & 65.8           & 66.3           & \underline{69.4}           & 65.7           & 62.4           & 48.9 & \underline{69.8}           & 69.9           \\
PatchCore(FPFH) & 59.7           & 57.4           & 65.4           & 68.7           & 51.2           & 52.4           & 53.1           & 62.5           & 32.7           & 72.0           & 35.8 & 45.9           & 57.1           \\
PatchCore(PointMAE) & 49.5           & 67.4           & 55.3           & 60.6           & 65.3           & 52.7           & 52.4           & 51.5           & 58.1           & 50.1           & 56.2  & 58.6           & 57.4           \\
CPMF            & 61.5           & 65.5           & 52.1           & 57.1           & 43.5           & 74.5           & 48.8           & 63.5           & 64.1           & 68.3           & 68.4  & 48.6           & 60.1           \\
Reg3D-AD        & \underline{69.8}           & 71.5           & \underline{88.6}           & 69.6           & 52.5           & 77.5           & 61.5           & 59.3           & 65.4           & \underline{80.0}           & 69.1  & 61.9           & 75.2           \\
IMRNet          & 67.1           & 66.8           & 55.6           & \underline{70.2}           & 64.1           & \underline{78.1}           & 70.5           & 68.4           & 59.9           & 57.6           & \underline{71.5} & 58.5           & \underline{77.4}           \\
PO3AD           & \textbf{96.2}           & \textbf{94.9}           & \textbf{91.2}           & \textbf{84.4}           & \textbf{88.0}           & \textbf{97.8}           & \textbf{91.4}           & \textbf{91.8}           & \textbf{93.5}           & \textbf{96.7}           & \textbf{94.1}  & \textbf{75.5}           & \textbf{89.9}           \\
\midrule
\multicolumn{14}{c}{\textit{Unified Method}}\\
\midrule
MC3D-AD      & \textbf{80.1} & \underline{81.5} & \underline{89.5} & \underline{88.3} & \underline{90.1} & \underline{82.7} & \underline{53.4} & \underline{60.7} & \underline{78.3} & \underline{65.4} & \underline{55.2} & \textbf{79.0} & \textbf{89.5}    \\ 
\rowcolor{gray!20}
\textbf{Ours}     & \underline{76.4} & \textbf{89.8} & \textbf{91.1} & \textbf{89.8} & \textbf{94.5} & \textbf{86.6} & \textbf{70.0} & \textbf{79.5} & \textbf{86.9} & \textbf{80.6} & \textbf{64.6} & \underline{71.7} & \underline{88.4} \\
\bottomrule
\end{tabular}
}

\resizebox{\textwidth}{!}{
\begin{tabular}{l|cccccccccccccc}
\toprule
Method & cap0 & cap3 & cap4 & cap5 & cup0 & cup1 & eraser0 & headset0 & headset1 & helmet0 & helmet1 & helmet2 & helmet3 \\
\midrule
\multicolumn{14}{c}{\textit{Category-Specific Method}}\\
\midrule
BTF(Raw)            & 52.4 &68.7 & 46.9 & 37.3 & 63.2 & 56.1 & 63.7 & 57.8 & 47.5 & 50.4 & 44.9 & 60.5 & 70.0 \\
BTF(FPFH)       & \underline{73.0}           & 65.8           & 52.4           & 58.6           & \underline{79.0}           & 61.9           & 71.9           & 62.0           & 59.1 & 57.5           & \underline{74.9}           & 64.3           & 72.4          \\
M3DM            & 53.1           & 60.5           & 71.8           & 65.5           & 71.5           & 55.6           & 71.0           & 58.1           & 58.5 & 59.9           & 42.7           & 62.3           & 65.5           \\
PatchCore(FPFH) & 47.2           & 65.3           & 59.5           & \underline{79.5}           & 65.5           & 59.6           & \underline{81.0}           & 58.3           & 46.4 & 54.8           & 48.9           & 45.5           & \underline{73.7}           \\
PatchCore(PointMAE) & 54.4           & 48.8           & 72.5           & 54.5           & 51.0           & \underline{85.6}           & 37.8           & 57.5           & 42.3 & 58.0           & 56.2           & 65.1           & 61.5           \\
CPMF            & 60.1           & 55.1           & 55.3           & 55.1           & 49.7           & 50.9           & 68.9           & 69.9           & 45.8 & 55.5           & 54.2           & 51.5           & 52.0           \\
Reg3D-AD        & 63.2           & \underline{71.8}           & \underline{81.5}           & 46.7           & 68.5           & 69.8           & 75.5           & 58.0           & \underline{62.6} & \underline{60.0}           & 62.4           & \underline{82.5}           & 62.0           \\
IMRNet          & 71.5           & 70.6           & 75.3           & 74.2           & 64.3           & 68.8           & 54.8           & \underline{70.5}           & 47.6  & 59.8           & 60.4           & 64.4           & 66.3          \\
PO3AD           & \textbf{95.7}           & \textbf{94.8}           & \textbf{94.0}           & \textbf{86.4}           & \textbf{90.9}           & \textbf{93.2}           & \textbf{97.4}           & \textbf{82.3}           & \textbf{90.7} & \textbf{87.8}           & \textbf{94.8}           & \textbf{93.2}           & \textbf{84.6}           \\
\midrule
\multicolumn{14}{c}{\textit{Unified Method}}\\
\midrule
MC3D-AD      & \underline{86.7} & \underline{92.2} & \underline{88.0} & \underline{88.5} & \underline{82.9} & \underline{71.5} & \textbf{85.6} & \underline{66.6} & \textbf{70.9} & \underline{74.4} & \underline{58.2} & \underline{81.7} & \underline{60.2}    \\ 
\rowcolor{gray!20}
\textbf{Ours}     & \textbf{90.0} & \textbf{97.2} & \textbf{94.0} & \textbf{95.0} & \textbf{87.1} & \textbf{84.2} & \underline{74.8} & \textbf{70.8} & \underline{68.7} & \textbf{81.0} & \textbf{59.6} & \textbf{88.5} & \textbf{70.1} \\
\bottomrule
\end{tabular}
}

\resizebox{\textwidth}{!}{
\begin{tabular}{l|cccccccccccccc|c}
\toprule
Method & jar & phone & shelf0 & tap0 & tap1 & vase0 & vase1 & vase2 & vase3 & vase4 & vase5 & vase7 & vase8 & vase9 & Mean \\
\midrule
\multicolumn{16}{c}{\textit{Category-Specific Method}}\\
\midrule
BTF(Raw)            & 42.3 & 58.3 & 46.4 & 52.7 & 56.4 & 61.8 & 54.9 & 40.3 & 60.2 & 61.3 & 58.5 & 57.8 & 55.0 & 56.4 & 55.0 \\
BTF(FPFH)        & 42.7           & 67.5           & 61.9           & 56.8           & 59.6           & 64.2           & 61.9 & 64.6           & \underline{69.9}           & 71.0           & 42.9           & 54.0           & 66.2           & 56.8           & 62.8            \\
M3DM            & 54.1           & 35.8           & 55.4           & 65.4           & 71.2           & 60.8           & 60.2 & 73.7           & 65.8           & 65.5           & 64.2           & 51.7           & 55.1           & 66.3           & 61.6            \\
PatchCore(FPFH) & 47.8           & 48.8           & 61.3           & 73.3           & \textbf{76.8}           & 65.5           & 45.3 & 72.1           & 43.0           & 50.5           & 44.7           & 69.3           & 57.5           & 66.3           & 58.0            \\
PatchCore(PointMAE) & 48.7           & \textbf{88.6}           & 54.3           & \textbf{85.8}           & 54.1           & \underline{67.7}           & 55.1 & \underline{74.2}           & 46.5           & 52.3           & 57.2           & 65.1           & 36.4           & 42.3           & 57.7            \\
CPMF            & 61.1           & 54.5           & \textbf{78.3}           & 45.8           & 65.7           & 45.8           & 48.6 & 58.2           & 58.2           & 51.4           & 65.1           & 50.4           & 52.9           & 54.5           & 57.3            \\
Reg3D-AD        & 59.9           & 59.9           & \underline{68.8}           & 58.9           & \underline{74.1}           & 54.8           & 60.2 & 40.5           & 51.1           & \underline{75.5}           & 62.4           & \underline{88.1}           & \underline{81.1}           & \underline{69.4}           & \underline{66.8}            \\
IMRNet          & \underline{76.5}           & 74.2           & 60.5           & 68.1           & 69.9           & 53.5           & \underline{68.5} & 61.4           & 40.1           & 52.4           & \underline{68.2}           & 59.3           & 63.5           & 69.1           & 65.0            \\
PO3AD           & \textbf{87.1}           & \underline{81.0}           & 66.3           & \underline{78.3}           & 69.2           & \textbf{95.5}           & \textbf{88.2} & \textbf{97.8}           & \textbf{88.4}           & \textbf{90.2}           & \textbf{93.7}           & \textbf{98.2}           & \textbf{95.0}           & \textbf{95.2}           & \textbf{89.8}           \\
\midrule
\multicolumn{16}{c}{\textit{Unified Method}}\\
\midrule
MC3D-AD       & \underline{84.8} & \underline{85.4} & \underline{65.8} & \underline{52.4} & \underline{53.7} & \textbf{87.0} & \textbf{72.6} & \underline{82.2} & \underline{79.7} & \underline{78.6} & \underline{64.7} & \underline{61.6} & \underline{88.5} & \underline{77.6} &  \underline{75.9}   \\ 
\rowcolor{gray!20}
\textbf{Ours}     & \textbf{91.1} & \textbf{89.1} & \textbf{67.9} & \textbf{59.7} & \textbf{63.0} & \underline{84.6} & \underline{72.2} & \textbf{85.0} & \textbf{83.0} & \textbf{88.4} & \textbf{76.6} & \textbf{77.7} & \textbf{90.0} & \textbf{83.5} & \textbf{81.0} \\
\bottomrule
\end{tabular}
}
\label{tab:asnet_category}
\end{table*}


\begin{table*}[t!]
\caption{Comparison of object-level AUPR (\%) of various methods on the Anomaly-ShapeNet dataset.}
\resizebox{\textwidth}{!}{
\begin{tabular}{l|ccccccccccccc}
\toprule
\multicolumn{14}{c}{O-AUPR($\uparrow$)}\\
\midrule
Method & ashtray0 & bag0 & bottle0 & bottle1 & bottle3 & bowl0 & bowl1 & bowl2 & bowl3 & bowl4 & bowl5 & bucket0 & bucket1 \\
\midrule
\multicolumn{14}{c}{\textit{Category-Specific Method}}\\
\midrule
    BTF(Raw)                 & 57.8  & 45.8  & 46.6  & 57.3  & 54.3  & 58.8  & 46.4  & 57.6  & \underline{65.4}  & 60.1  & 61.5  & 65.2  & 62.0 \\
BTF(FPFH)                & 65.1  & 55.1  & 64.4  & 62.5  & 60.2  & 57.6  & \underline{64.8}  & 51.5  & 49.9  & 63.2  & \underline{69.9}  & 48.3  & 64.8 \\
M3DM                     & 63.2  & 64.2  & \underline{76.3}  & 67.4  & 45.1  & 52.5  & 51.5  & 63.0  & 63.5  & 57.1  & 60.1  & 60.9  & 50.7 \\
PatchCore(FPFH)          & 44.5  & 60.8  & 61.5  & 67.7  & 57.9  & 54.8  & 54.5  & 61.1  & 62.0  & 57.5  & 54.1  & 60.4  & 56.5 \\
PatchCore(PointMAE)      & \underline{67.9}  & 60.1  & 54.5  & 64.5  & \underline{65.1}  & 56.2  & 61.1  & 45.6  & 55.6  & 60.1  & 58.5  & 54.1  & 64.2 \\
CPMF                     & 45.3  & 65.5  & 58.8  & 59.2  & 50.5  & \underline{77.5}  & 62.1  & 60.1  & 41.8  & \underline{68.3}  & 68.5  & \underline{66.2}  & 50.1 \\
Reg3D-AD                 & 58.8  & 60.8  & 63.2  & 69.5  & 47.4  & 49.4  & 51.5  & 49.5  & 44.1  & 62.4  & 55.5  & 63.2  & 71.4 \\
IMRNet                   & 61.2  & \underline{66.5}  & 55.8  & \underline{70.2}  & 64.8  & 48.1  & 50.4  & \underline{68.1}  & 61.4  & 63.0  & 65.2  & 57.8  & \underline{73.2} \\
PO3AD                    & \textbf{99.9}  & \textbf{80.9}  & \textbf{92.7}  & \textbf{95.9}  & \textbf{96.2}  & \textbf{94.6}  & \textbf{90.5}  & \textbf{88.8}  & \textbf{92.7}  & \textbf{98.5}  & \textbf{90.4}  & \textbf{92.3}  & \textbf{88.2} \\

\midrule
\multicolumn{14}{c}{\textit{Unified Method}}\\
\midrule
\rowcolor{gray!20}
\textbf{Ours} & \textbf{97.5} & \textbf{88.0} & \textbf{93.4} & \textbf{94.3} & \textbf{99.0} & \textbf{97.5} & \textbf{96.1} & \textbf{93.6} & \textbf{97.0} & \textbf{98.9} & \textbf{98.5} & \textbf{94.1} & \textbf{94.4} \\
\bottomrule
\end{tabular}
}

\resizebox{\textwidth}{!}{
\begin{tabular}{l|cccccccccccccc}
\toprule
Method & cap0 & cap3 & cap4 & cap5 & cup0 & cup1 & eraser0 & headset0 & headset1 & helmet0 & helmet1 & helmet2 & helmet3 \\
\midrule
\multicolumn{14}{c}{\textit{Category-Specific Method}}\\
\midrule
BTF(Raw)            & 65.9  & 61.2  & 51.5  & 65.3  & 60.1  & 70.1  & 42.5  & 37.9  & 51.5  & 55.9  & 38.8  & 61.5  & 52.6 \\
BTF(FPFH)           & 61.8  & 57.9  & 54.5  & 59.3  & 58.5  & 65.1  & 71.9  & 53.1  & 52.3  & 56.8  & \underline{72.1}  & 58.8  & 56.4 \\
M3DM                & 56.4  & 65.2  & 47.7  & 64.2  & 57.0  & \underline{75.2}  & 62.5  & 63.2  & 62.3  & 52.8  & 62.7  & \underline{63.6}  & 45.8 \\
PatchCore(FPFH)     & 58.5  & 45.7  & 65.5  & 72.5  & 60.4  & 58.6  & 58.4  & \underline{70.1}  & 60.1  & 52.5  & 63.0  & 47.5  & 49.4 \\
PatchCore(PointMAE) & 56.1  & 58.3  & \underline{72.1}  & 54.2  & 64.2  & 71.0  & \underline{80.1}  & 51.5  & 42.3  & 63.3  & 57.1  & 49.6  & 61.1 \\
CPMF                & 60.1  & 54.1  & 64.5  & 69.7  & \underline{64.7}  & 60.9  & 54.4  & 60.2  & 61.9  & 33.3  & 50.1  & 47.7  & \underline{64.5} \\
Reg3D-AD            & 69.3  & \underline{71.1}  & 62.3  & \underline{77.0}  & 53.1  & 63.8  & 42.4  & 53.8  & 61.7  & 60.0  & 38.1  & 61.8  & 46.8 \\
IMRNet              & \underline{71.1}  & 70.2  & 65.8  & 50.2  & 45.5  & 62.7  & 59.9  & \underline{70.1}  & \underline{65.6}  & \underline{69.7}  & 61.5  & 60.2  & 57.5 \\
PO3AD               & \textbf{84.1}  & \textbf{90.6}  & \textbf{87.6}  & \textbf{80.1}  & \textbf{87.9}  & \textbf{87.0}  & \textbf{99.5}  & \textbf{76.5}  & \textbf{91.4}  & \textbf{86.4}  & \textbf{96.1}  & \textbf{93.4}  & \textbf{84.9} \\

\midrule
\multicolumn{14}{c}{\textit{Unified Method}}\\
\midrule
\rowcolor{gray!20}
\textbf{Ours}   & \textbf{98.8} & \textbf{99.2} & \textbf{99.7} & \textbf{87.5} & \textbf{99.5} & \textbf{100.0} & \textbf{57.8} & \textbf{86.9} & \textbf{89.9} & \textbf{97.9} & \textbf{100.0} & \textbf{89.4} & \textbf{100.0} \\
\bottomrule
\end{tabular}
}

\resizebox{\textwidth}{!}{
\begin{tabular}{l|cccccccccccccc|c}
\toprule
Method & jar & phone & shelf0 & tap0 & tap1 & vase0 & vase1 & vase2 & vase3 & vase4 & vase5 & vase7 & vase8 & vase9 & Mean \\
\midrule
\multicolumn{16}{c}{\textit{Category-Specific Method}}\\
\midrule
BTF(Raw)            & 42.8  & 61.3 & 62.4  & 53.5  & 59.4  & 56.2  & 44.1  & 41.3  & \underline{71.7}  & 42.8  & 61.5  & 54.7  & 41.6  & 48.2  & 54.9 \\
BTF(FPFH)           & 47.9  & \underline{66.2} & 61.1  & 61.0  & 57.5  & 64.1  & 65.5  & 56.9  & 65.2  & 58.7  & 47.2  & 59.2  & 62.4  & 63.8  & 59.8 \\
M3DM                & 55.5  & 46.4 & 66.5  & \underline{72.2}  & 63.8  & \textbf{78.8}  & 65.2  & 61.5  & 55.1  & 52.6  & 63.3  & 64.8  & 46.3  & 65.1  & 60.3 \\
PatchCore(FPFH)     & 49.9  & 33.2 & 50.4  & 71.2  & 68.4  & 64.5  & 62.3  & \underline{80.1}  & 48.1  & \underline{77.7}  & 51.5  & 62.1  & 51.5  & \underline{66.0}  & 58.8 \\
PatchCore(PointMAE) & 46.3  & 65.2 & 54.3  & 71.2  & 54.2  & 54.8  & 57.2  & 71.1  & 45.5  & 58.6  & 58.5  & \underline{65.2}  & 65.5  & 63.4  & 59.5 \\
CPMF                & 61.8  & 65.5 & \textbf{68.1}  & 63.9  & 69.7  & 63.2  & 64.5  & 63.2  & 58.8  & 65.5  & 51.8  & 43.2  & \underline{67.3}  & 61.8  & 59.7 \\
Reg3D-AD            & 60.1  & 61.4 & 67.5  & 67.6  & 59.9  & 61.5  & 46.8  & 64.1  & 65.1  & 50.5  & 58.8  & 45.5  & 62.9  & 57.4  & 58.4 \\
IMRNet              & \underline{76.0}  & 55.2 & 62.5  & 40.1  & \textbf{79.6}  & 57.3  & \underline{72.5}  & 65.5  & 70.8  & 52.8  & \underline{65.4}  & 60.1  & 63.9  & 46.2  & \underline{62.1} \\
PO3AD               & \textbf{91.5}  & \textbf{80.3} & \underline{68.0}  & \textbf{85.6}  & \underline{70.9}  & \underline{75.3}  & \textbf{78.9}  & \textbf{96.3}  & \textbf{90.2}  & \textbf{82.4}  & \textbf{87.9}  & \textbf{97.1}  & \textbf{83.3}  & \textbf{90.4}  & \textbf{88.1} \\

\midrule
\multicolumn{16}{c}{\textit{Unified Method}}\\
\midrule
\rowcolor{gray!20}
\textbf{Ours}   & \textbf{100.0} & \textbf{88.8} & \textbf{91.7} & \textbf{93.9} & \textbf{82.6} & \textbf{87.6} & \textbf{97.4} & \textbf{97.5} & \textbf{84.7} & \textbf{95.3} & \textbf{94.8} & \textbf{100.0} & \textbf{91.0} & \textbf{96.4} & \textbf{93.8} \\
\bottomrule
\end{tabular}
}

\label{tab:o_aupr_as}
\end{table*}

\begin{figure*}[!t] 
\centerline{\includegraphics[width=0.75\linewidth]{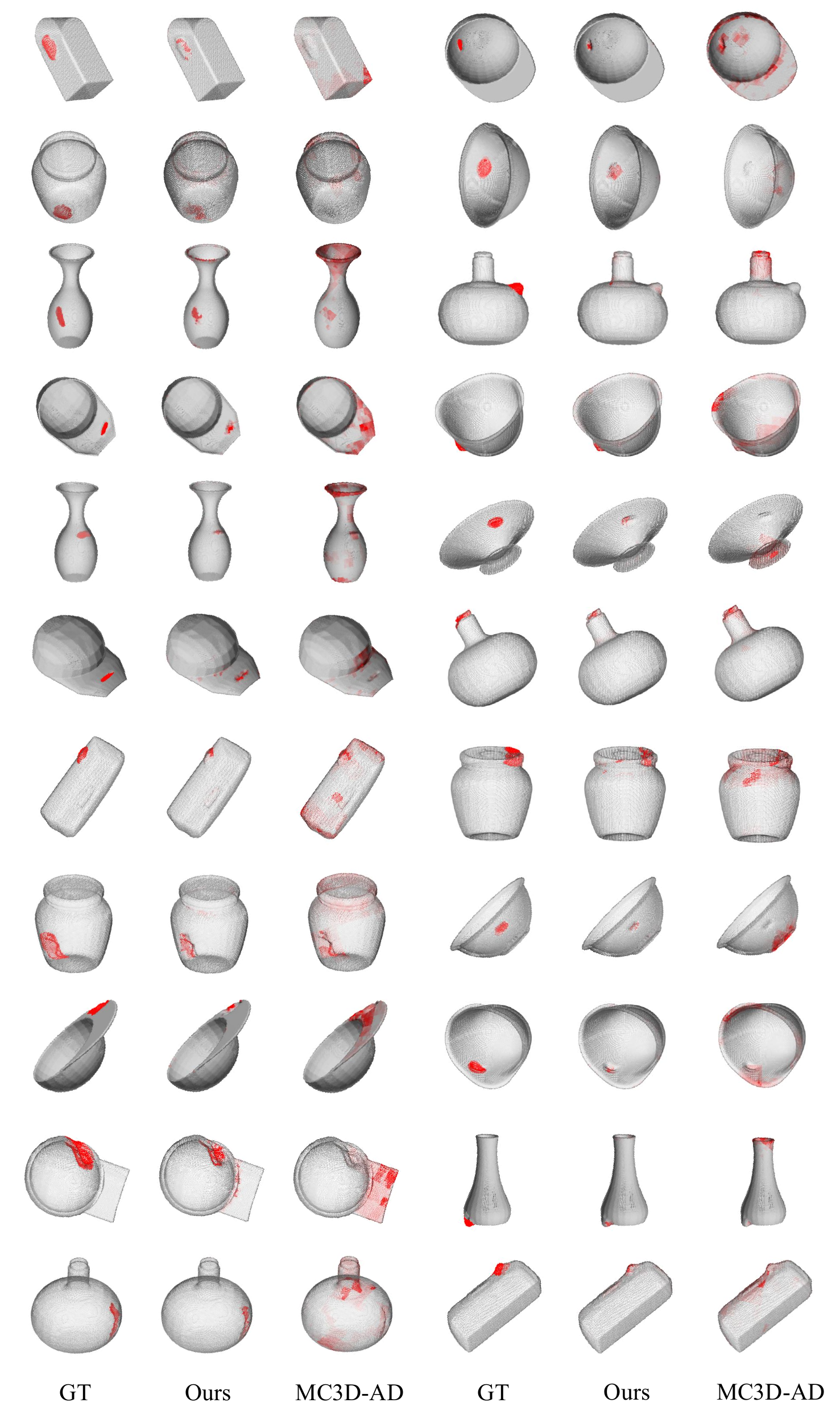}}
    \caption{Additional qualitative comparison with MC3D-AD on the Anomaly-ShapeNet dataset.}
    \label{fig:supple_fig2} 
\end{figure*}

\renewcommand{\arraystretch}{1.08}

\end{document}